\documentclass[conference]{IEEEtran}
\usepackage{times}

\usepackage[numbers]{natbib}
\usepackage{multicol}
\usepackage[bookmarks=true]{hyperref}
\usepackage{makecell}
\usepackage{xcolor}
\usepackage{cuted}
\usepackage{caption}
\usepackage{amsmath}
\usepackage{amssymb}
\usepackage{mathtools}
\usepackage{amsthm}
\usepackage{multirow}
\usepackage{booktabs}
\usepackage{adjustbox}
\usepackage{amssymb}
\usepackage{bm}
\usepackage{algorithm}
\usepackage{algorithmic}
\usepackage[table]{xcolor}

\usepackage{siunitx}
\newcommand{\authnote}[2][]{%
  \ifx&#1&%
    $\ll$\textsf{\footnotesize #2}$\gg$
  \else
    $\ll$\textsf{\footnotesize #1$\Vert$#2}$\gg$
  \fi
}

\newcommand{\greencheck}{{\color{black}$\checkmark$}} %
\newcommand{\redcross}{{\color{black}$\times$}} %

\definecolor{oursblue}{RGB}{40,90,180}
\newcommand{\ours}{\textbf{\textcolor{oursblue}{TextOp}}}

\begin{document}

\title{TextOp: Real-time Interactive Text-Driven \\Humanoid Robot Motion Generation and Control}

\author{%
  Weiji Xie\textsuperscript{1,2*} \quad
  Jiakun Zheng\textsuperscript{1,3*} \quad
  Jinrui Han\textsuperscript{1,2} \quad
  Jiyuan Shi\textsuperscript{1}\\
  Weinan Zhang\textsuperscript{2\dag{}}\quad 
  Chenjia Bai\textsuperscript{1\dag{}}\quad 
  Xuelong Li\textsuperscript{1}\\
  \textsuperscript{1}Institute of Artificial Intelligence (TeleAI), China Telecom \quad
  \textsuperscript{2}Shanghai Jiao Tong University \\
  \textsuperscript{3}East China University of Science and Technology \\
  \textsuperscript{*}Equal contribution \quad
  \textsuperscript{\dag}Corresponding author\\
}

\maketitle

\begin{strip}
  \centering
  \includegraphics[width=0.99 \linewidth]{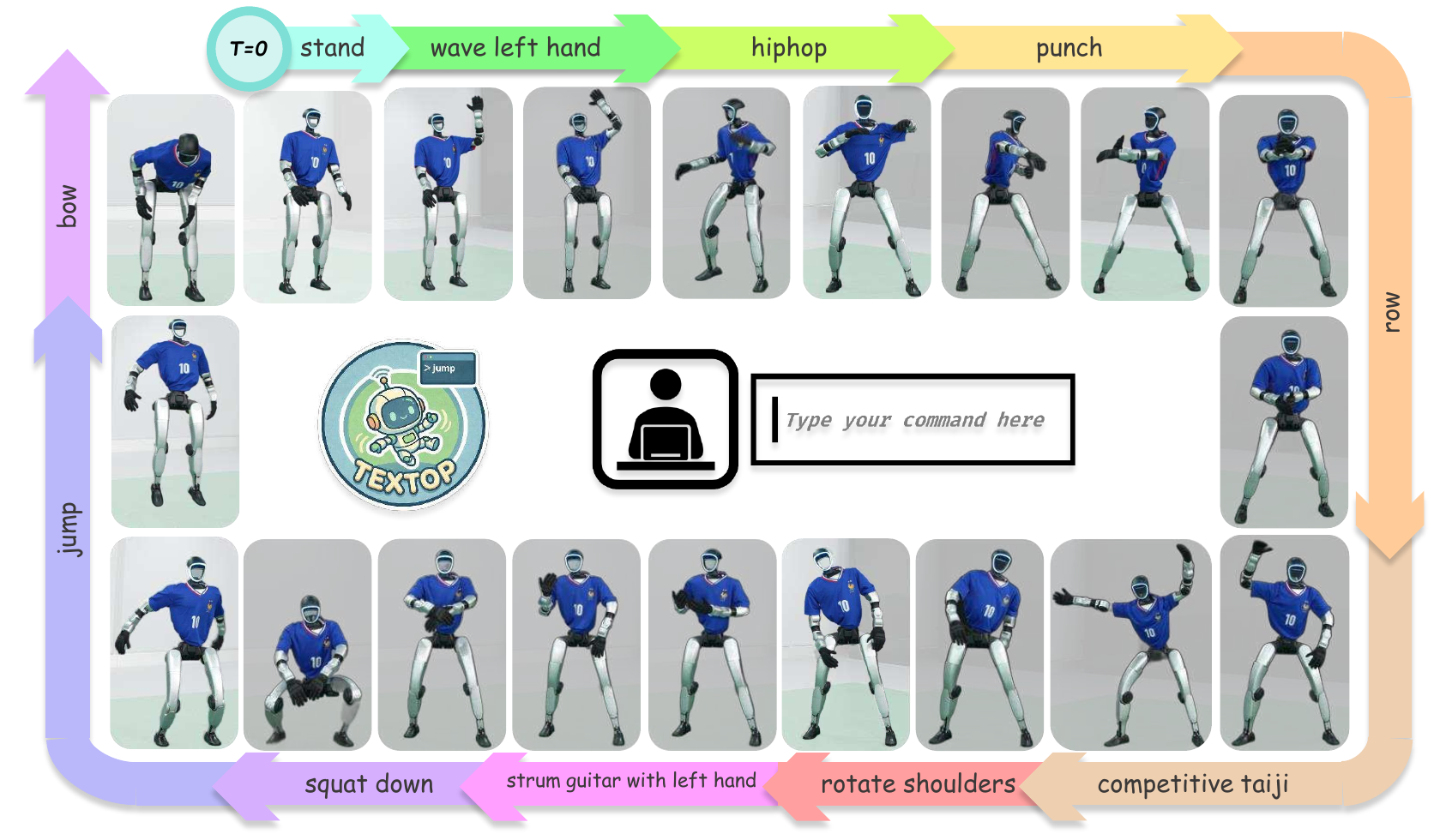}
  
  \captionof{figure}{\textbf{\ours{} enables a humanoid robot to execute a seamless sequence of diverse skills—ranging from expressive gestures to complex physical tasks—driven by real-time, interactive text commands from the user in a single continuous trial.} }
  \label{fig:teaser}
\end{strip}

\begin{abstract}
Recent advances in humanoid whole-body motion tracking have enabled the execution of diverse and highly coordinated motions on real hardware. 
However, existing controllers are commonly driven either by predefined motion trajectories, which offer limited flexibility when user intent changes, or by continuous human teleoperation, which requires constant human involvement and limits autonomy.
This work addresses the problem of how to drive a universal humanoid controller in a real-time and interactive manner. We present \ours{}, a real-time text-driven humanoid motion generation and control framework that supports streaming language commands and on-the-fly instruction modification during execution. \ours{} adopts a two-level architecture in which a high-level autoregressive motion diffusion model continuously generates short-horizon kinematic trajectories conditioned on the current text input, while a low-level motion tracking policy executes these trajectories on a physical humanoid robot. By bridging interactive motion generation with robust whole-body control, \ours{} unlocks free-form intent expression and enables smooth transitions across multiple challenging behaviors such as dancing and jumping, within a single continuous motion execution. Extensive real-robot experiments and offline evaluations demonstrate instant responsiveness, smooth whole-body motion, and precise control. The project page and the open-source code are available at \url{https://text-op.github.io/}.

\end{abstract}

\IEEEpeerreviewmaketitle

\section{Introduction}
\label{sec:intro}

\begin{figure*}[t]
  \centering
  \includegraphics[width=0.9 \linewidth]{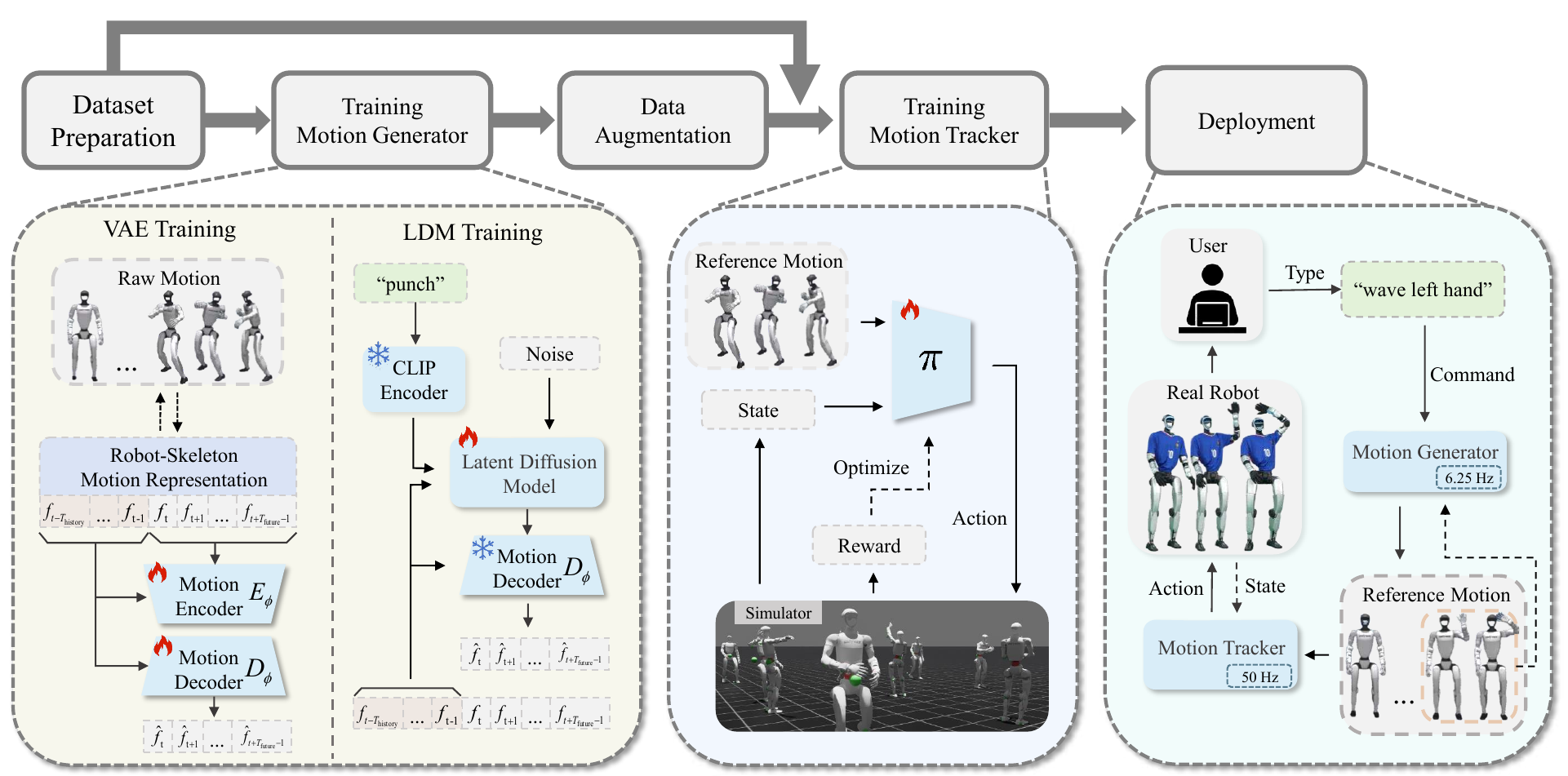}
  \caption{\textbf{Overview of \ours{}'s framework.} 
    The framework consists of three main parts: 
    (a) \textbf{Interactive Motion Generation}, including VAE training and LDM training, which together model future reference motion sequences conditioned on history motion and text prompt in an autoregressive style; 
    (b) \textbf{Dynamic Motion Tracking}, where the MLP-based policy $\pi$ takes reference motions and robot states to generate joint actions, trained in the simulation for stable execution; 
    (c) \textbf{Deployment}, where the real-time user text prompt is converted into motions by the generator, translated into actions by the tracking policy based on the robot state, and executed on the physical robot.}
  \label{fig:method}
\end{figure*}

Recent years have witnessed rapid progress in universal humanoid whole-body motion tracking systems~\cite{luo2025sonic,he2024omnih2o,chen2025gmt,han2025kungfubot2,yin2025unitracker,li2025bfm0}, capable of robustly executing a wide range of highly coordinated behaviors on real hardware. Leveraging advances in reinforcement learning~\cite{barto2021reinforcement,schulman2017ppo} and large-scale physical simulation~\cite{mittal2025isaaclab,todorov2012mujoco}, modern humanoid platforms can reliably track diverse kinematic references and perform complex full-body motions under a unified control framework. From a control perspective, the problem of executing rich humanoid behaviors is increasingly well-addressed.

However, \textbf{how to flexibly and interactively drive such universal controllers remains an open challenge}. In existing systems, universal humanoid controllers are commonly driven by either predefined motion trajectories~\cite{chen2025gmt,han2025kungfubot2,zhang2025any2track} or continuous human teleoperation~\cite{ze2025twist,ze2025twist2,he2024omnih2o}. Predefined trajectories, while effective for reproducing specific motions, offer little flexibility at runtime: once execution begins, the robot’s behavior is mainly determined by the trajectory and cannot easily adapt to changes in intent. Teleoperation, on the other hand, enables direct guidance but relies heavily on sustained human involvement, making long-term operation labor-intensive and limiting the autonomy of the robot. As a result, a gap persists between high-level human intent expression and low-level humanoid execution, despite significant advances in the underlying control technology.

In this context, natural language has emerged as a powerful modality for guiding robot behavior. A growing body of work on text-conditioned motion generation demonstrates that language can flexibly describe complex, semantically meaningful motions~\cite{chen2023mld,tevet2022mdm,jiang2023motiongpt}. Inspired by this idea, some approaches first explicitly generate full motion sequences and then employ a tracking policy to track them on the robot~\cite{he2024omnih2o,serifi2024robotmdm,yin2025unitracker,han2025kungfubot2}, while others embed motion generation implicitly within end-to-end models that map language directly to control signals~\cite{shao2025langwbc,wang2025sentinel,li2025roboghost}. Although differing in formulation, these methods share a common characteristic: behavior is determined largely upfront, with limited or no support for revising commands during execution. This limits their applicability to real-world humanoid control, where interaction is continuous and intent may evolve over time.

Separately, the humanoid animation community has explored interactive motion generation techniques that allow users to steer motions on the fly~\cite{chen2024camdm,shi2024amdm,zhao2024dartcontrol,xiao2025motionstreamer,cai2025flooddiffusion}. 
A key characteristic is that both the conditioning signal and the generated motion are treated as time-varying signal streams, rather than producing a complete trajectory from a static prompt.
These approaches show the feasibility of responsive, user-in-the-loop motion synthesis, but are primarily developed for kinematic animation and virtual characters, without integration into real-robot whole-body control pipelines, due to missing dynamic constraints and the unaddressed sim-to-real gap. 

Taken together, these observations reveal a missing link between interactive language-based intent expression and real-time, physically executable humanoid control.
Based on this insight, we propose \ours{}, a real-time text-driven humanoid motion generation and control framework that drives a universal whole-body controller using streaming language commands with on-the-fly instruction modification. Rather than committing to a fixed motion plan, \ours{} continuously interprets evolving text input and generates short-horizon motion references suitable for immediate execution on real hardware.

\ours{} adopts a two-level architecture. At the high level, an autoregressive, text-conditioned motion diffusion model incrementally generates short-horizon kinematic trajectories conditioned on the current language input and recent motion context. At the low level, a robust whole-body motion tracking policy executes these trajectories on a physical humanoid robot, converting kinematic references into joint-level commands at high frequency. This separation allows human intent to be updated flexibly without compromising control stability.

To effectively integrate interactive motion generation with real-world humanoid control, \ours{} incorporates two critical design choices tailored to robotic execution. First, we adopt a robot-skeleton motion representation that reflects the constrained, single-DoF joint structure of humanoid robots, enabling compact and effective kinematic motion representation for the motion generation module. Second, to mitigate the distribution gap between motion datasets and the reference motions produced online by the generator, we augment the training data of the tracking policy with generator-produced motions under realistic text streams. Together, these designs align the high-level generator with the low-level controller and improve robustness during real-robot deployment.

By bridging interactive motion generation with whole-body control, \ours{} enables the robot to execute long-horizon, continuous motions that smoothly transition across multiple challenging behaviors—such as dancing and jumping—while dynamically responding to user-provided language commands in real time, as validated on a real robot. Offline evaluations further demonstrate that carefully designed robot-skeleton motion representations, together with augmenting tracker training with generator-produced motions, enable the system to reliably translate streaming language commands into real-world robot motion.

In summary, this work makes the following contributions:
\begin{itemize}
    \item We present \textbf{\ours{}}, a real-time text-driven humanoid motion generation and control system that realizes a novel control mode in which the robot is driven by streaming natural language commands.
    
    \item We introduce two key design choices: (i) a robot-skeleton motion representation to improve generation quality for robot structure, and (ii) a data-augmentation strategy via motion generation to reduce the distribution gap between motion datasets and generator-produced reference motions.
    
    \item We validate \ours{} through extensive real-robot experiments and offline evaluations, demonstrating instant responsiveness, smooth whole-body motion, and precise control.
\end{itemize}

\section{Related Work}
\label{sec:related_work}

\subsection{Whole-body Humanoid Control}
\label{sec:related_work-wbc}
Humanoid whole-body control has advanced significantly, enabling realistic, diverse, and coordinated motion execution on physical robots. From the perspective of the driving mechanism, humanoid robot controllers are mainly driven by three command paradigms.

\textbf{Simple Task Command} encompasses control methods that enable the robot to execute straightforward high-level objectives, such as locomotion~\cite{gu2024humanoidgym,escontrela2025gaussgym,zhu2026hiking,wang2025more,xue2025hugwbc,xie2025dbhl}, navigation~\cite{ben2025gallant,cai2025navdp,chen2025head}, object manipulation~\cite{he2025viral,xue2025doorman}, sports~\cite{su2025hitter,haarnoja2024soccer,ren2025goalkeeper}, and standing up~\cite{huang2025host,he2025humanup}. This paradigm often provides fast response and reliability for single tasks, but it offers limited expressiveness for complex, semantically rich motions.

\textbf{Static motion tracking} relies on predefined reference trajectories to achieve agile and high-fidelity whole-body motion reproduction, but commits to a fixed motion plan that is difficult to adjust in real time. This paradigm leverages motion data from motion capture~\cite{mahmood2019amass,hwang2025snapmogen}, video-based extraction~\cite{allshire2025videomimic,shen2024gvhmr,wang2025prompthmr}, and multimodal generation models in both explicit~\cite{he2024omnih2o,serifi2024robotmdm} and implicit~\cite{shao2025langwbc,wang2025sentinel,li2025roboghost,xue2025leverb} ways, together with motion retargeting~\cite{araujo2025gmr,yang2025omniretarget} and reinforcement learning methods~\cite{liao2025beyondmimic,luo2025sonic,xie2025kungfubot,han2025kungfubot2,zhang2025any2track,he2024omnih2o,chen2025gmt,pan2025ams,yin2025unitracker,chen2025chip,zhuang2026deepwholebodyparkour} for whole-body motion tracking. 
We note that a few prior methods~\cite{shao2025langwbc,luo2025sonic} technically allow modifying text inputs at runtime; however, this capability is not a design focus and is only weakly supported in practice, with no systematic empirical validation.

\textbf{Teleoperation} provides flexible real-time control through motion capture suits~\cite{ze2025twist}, VR devices~\cite{ze2025twist2,lu2025mobiletelevision}, camera-based systems~\cite{he2024h2o}, or exoskeletons~\cite{myers2025child,yang2025ace,ben2025homie}, but requires continuous human involvement, limiting autonomy and scalability for long-term operation.

This work goes beyond both static motion tracking and teleoperation paradigms, achieving flexible semantic intent expression and continuous real-time control with little reliance on manual labor.

\subsection{Interactive Motion Generation}
\label{sec:related_work-mg}
Text-driven motion generation has advanced significantly, with natural language emerging as a powerful modality for describing complex, semantically rich motions. Diffusion-based approaches~\cite{zhang2024motiondiffuse,tevet2022mdm,chen2023mld}, followed by language-model-integrated methods~\cite{zhang2023t2mgpt, jiang2023motiongpt, wang2024motiongpt, zhu2025motiongpt3}, have demonstrated that compact textual descriptions can be effectively translated into diverse and high-quality human motions. Nevertheless, most existing motion generation systems remain offline, producing complete motion sequences from static inputs.

Interactive motion generation has been proposed to address this limitation. Early approaches enable interactive control using low-dimensional signals such as target position and direction~\cite{chen2024camdm, shi2024amdm}, while later work incorporate text as control signals.
\citep{zhao2024dartcontrol} introduces a diffusion-based autoregressive motion primitive model that conditions on historical frames and text prompts for real-time generation. \citep{zhang2025primal} uses a two-stage method that first pretrain an autoregressive diffusion model to capture short-term body movements and then adapts it for real-time interactive control.
\citep{xiao2025motionstreamer} performs continuous causal latent autoregression, alleviating error accumulation caused by discrete tokenization.
\citep{cai2025flooddiffusion} proposes a diffusion-forcing framework with tailored temporal attention and scheduling to enable seamless real-time synthesis.

Despite these advances, existing methods focus on virtual character animation and lack physical robot deployment.
Our work bridges this gap by integrating autoregressive motion generation with a whole-body tracking policy for real humanoid control.

\section{Method}
\label{sec:method}

\begin{figure}[t]
  \centering
  \includegraphics[width=1 \linewidth]{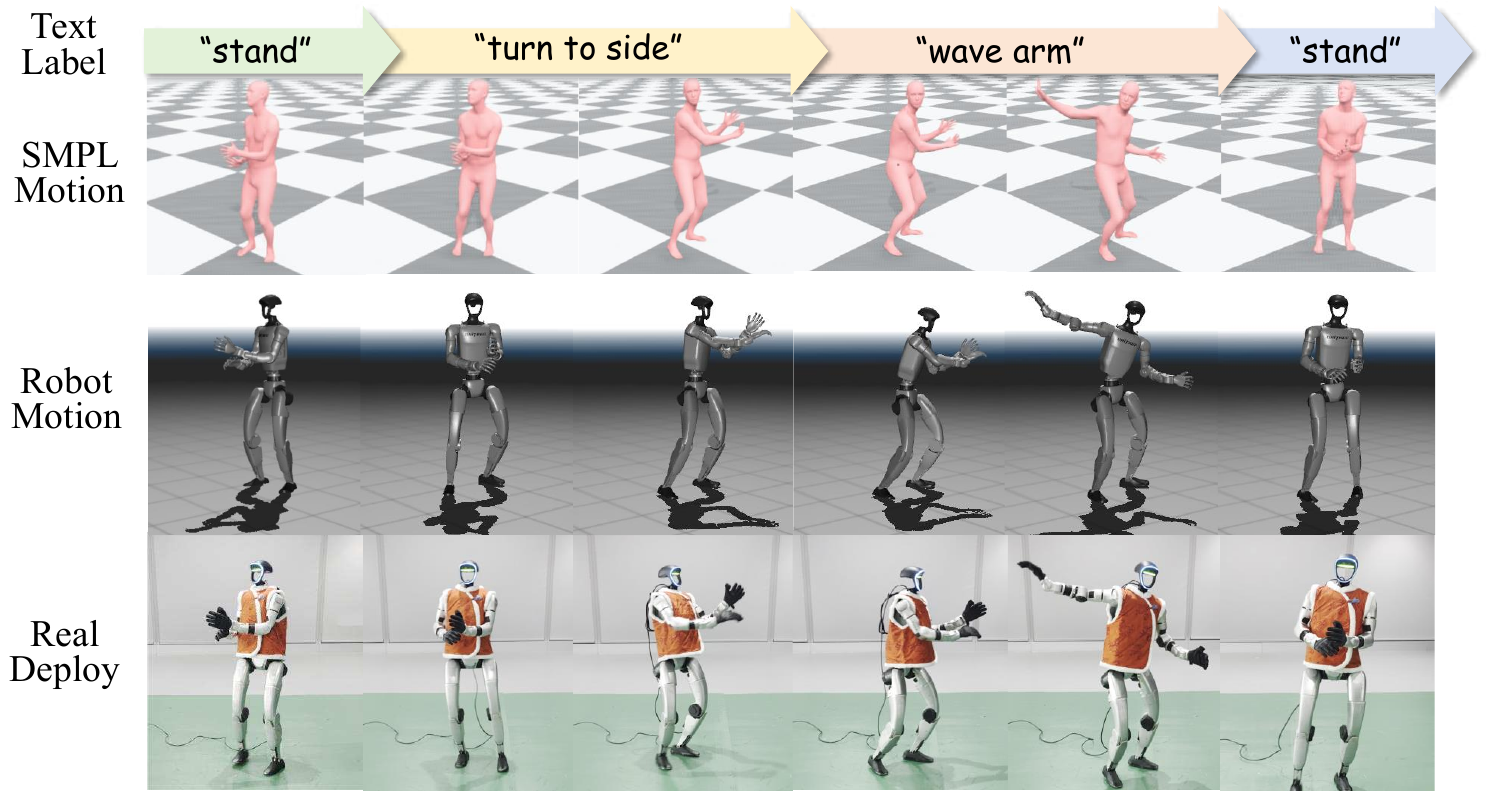}
  \caption{\textbf{Illustration of the time-aligned data format including text labels, SMPL motions, and robot motions.}}
  \label{fig:timeline}
\end{figure}

\subsection{Overview}
\label{sec:method-overview}

We consider a humanoid robot operating in a real-time streaming control setting. At each discrete time step $t$, the system receives a language command $l_t$ together with the previous robot state $x^\text{robot}_{<t}$. The goal of the system is to generate a control signal, i.e., target joint commands $a_t$, such that the resulting physical robot behavior $x^\text{robot}_t$ remains semantically consistent with the language command while maintaining stable and smooth whole-body motion.

\ours{} decomposes this task into two levels: Interactive Motion Generation and Dynamic Motion Tracking. Specifically, the high-level motion generator $G$ is implemented as an autoregressive motion latent diffusion model. At each time step, it produces a reference motion sequence $x_{t:t+T_\text{future}-1}^\text{ref}$ of length $T_\text{future}=8$,  conditioned on the past $T_\text{history}=2$ frames of reference motion $x_{t-T_\text{history}:t-1}^\text{ref}$ and the current language command $l_t$, i.e.,
\[
x_{t:t+T_\text{future}-1}^\text{ref} = G\big(x_{t-T_\text{history}:t-1}^\text{ref}, l_t\big).
\]
The low-level tracking policy $\pi$ then converts the kinematic reference motion into executable control signals based on the previous robot states, i.e.,
\[
a_t = \pi\big(x_{t-1}^\text{robot},a_{t-1}, x_{t:t+T_\text{ref}-1}^\text{ref}\big),
\]
ensuring that the executed motion closely follows the reference trajectory while remaining stable and smooth.

\subsection{Data Preparation}
\label{sec:method-data}

Our training data is built on the large-scale public motion capture dataset AMASS~\cite{mahmood2019amass}, supplemented with a small amount of private motion data featuring complex dance and martial arts motions. All motions are retargeted from the SMPL~\cite{loper2023smpl} skeleton to the robot skeleton using GMR~\cite{araujo2025gmr}, extracting the feet contact indicator and resampling to a unified frequency of 50 Hz. We further filter the retargeted motions by pretraining a universal motion tracking policy with privileged information and discarding motion sequences that cannot be reliably tracked. Fig. \ref{fig:timeline} illustrates our data format across different modalities.

For training the motion tracking policy, the filtered motions are further segmented into short clips of less than 40 seconds, resulting in a total of 12,296 motion clips with an aggregate duration of 40.67 hours from AMASS. The private dataset contributes an additional 403 clips totalling 3.12 hours.

For motion generation, we pair filtered AMASS motions with language annotations from BABEL~\cite{babel}, an action-centric dataset in which long motion sequences are segmented, each annotated with a text description, e.g., ``walk'', ``jump''. Such frame-level annotations are essential for effective language guidance and natural switching during generation, as demonstrated in prior work~\cite{zhao2024dartcontrol,cai2025flooddiffusion}. We perform mirror augmentation on both the motion sequences and their corresponding textual descriptions, resulting in a dataset of 83,478 segment–text pairs. For complex long-horizon motions in the private dataset, the entire motion is assigned a unique label wrapped with ``$\langle \cdot \rangle$'' markers.

\subsection{Interactive Motion Generation}
\label{sec:method-mg}

\textbf{Robot-Skeleton Motion Representation.} In contrast to humanoid skeletons like SMPL, where joints are modeled as 3-DoF ball joints to represent full human articulation, robotic skeletons typically employ single-DoF rotational joints, yielding more constrained yet tractable kinematics. Therefore, we propose a DoF-based local incremental motion representation for the robot skeleton, instead of HumanML3D \cite{HumanML3D}-style representation. This choice naturally enforces robot kinematic constraints and ensures invariance to global pose, providing a compact yet sufficient feature space for the motion generator.

For a robot with $n_q$ degrees of freedom, we define the per-frame motion feature $f_t \in \mathbb{R}^{d_\text{feat}}$ as:
\[
f_t =
\Big[
\phi(r_t),\;
\Delta \psi_t,\;
c_t,\;
\Delta p_t^{\text{local}},\;
h_t,\;
q_t,\;
\Delta q_t
\Big],
\]
where $r_t = (\text{roll}_t, \text{pitch}_t, \text{yaw}_t)$ denotes the root orientation in intrinsic Euler angles and $\phi(r_t) = [\sin(\text{roll}_t), \cos(\text{roll}_t)-1, \sin(\text{pitch}_t), \cos(\text{pitch}_t)-1]$ is a continuous trigonometric encoding of roll and pitch. $\Delta \psi_t = \text{yaw}_{t+1} - \text{yaw}_t$ denotes the incremental change of the root yaw, $c_t \in \{0,1\}^{n_c}$ represents binary contact indicators of feet, and $\Delta p_t^{\text{local}} = R_z(\text{yaw}_t)^\top (p_{t+1} - p_t)$ is the root translation increment expressed in a yaw-aligned local frame. Furthermore, $h_t$ denotes the root height, $q_t \in \mathbb{R}^{n_q}$ the joint positions, and $\Delta q_t = q_{t+1} - q_t$ the corresponding joint-wise increments.

\textbf{Model Architecture.} 
The motion generator $G$ models the text-conditional probabilistic distribution of the robot motion in an autoregressive style. It follows the DART~\cite{zhao2024dartcontrol} architecture and is implemented as a VAE combined with a latent diffusion model (LDM), both are Transformer-based networks.
The VAE defines a motion latent space by encoding future motion frames into a latent variable conditioned on the motion history, $z \sim E_\phi(\cdot \mid f_{t-T_\text{history}:t+T_\text{future}-1})$, which can be decoded to reconstruct the future frames $\hat{f}_{t:t+T_\text{future}-1} = D_\phi(f_{t-T_\text{history}:t-1}, z)$. 
The LDM models the text-conditional distribution of future motion latents, where language commands are embedded using a pretrained CLIP encoder~\cite{radford2021clip} as
$e_t = \text{CLIP}(l_t)$, and a diffusion transformer predicts clean latents from noise via
$\hat{z} = F_\theta(z_k, k, f_{t-T_\text{history}:t-1}, e_t)$,
trained in a DDPM~\cite{ho2020ddpm} style with $n_\text{denoise}=5$ steps.
During inference, classifier-free guidance is applied with scale $\sigma_\text{CFG}=5$ to enhance semantic alignment.

\textbf{Training Process.}
The VAE is first trained on motion features reconstruction, and its parameters are fixed during subsequent LDM training. The LDM is then trained to denoise motion latents by minimizing a combination of feature reconstruction, latent reconstruction, and geometric losses. To reduce the distribution gap between training and deployment, a self-rollout strategy is adopted, where $N$ consecutive overlapping motion segments are processed sequentially and the history of each segment is randomly replaced by the predicted future of the previous segment. For classifier-free guidance, the text embedding is randomly set to zero during training.

\subsection{Dynamic Motion Tracking}
\label{sec:method-wbc}

\textbf{Training Universal Motion Tracker.} 
The low-level universal motion tracker is trained through goal-conditioned RL in large-scale physical simulation with IsaacLab~\cite{mittal2025isaaclab}. We adopt a one-stage RL pipeline to train the MLP-based tracking policy, discarding more complex training pipelines and network designs to maintain simplicity while ensuring balanced performance.

The policy takes as input the robot’s proprioceptive state together with a reference motion over $T_\text{ref}=5$ future frames. During training, an episode starts with a motion segment randomly sampled from the dataset and the robot initialized to the state of the first frame, and ends upon timeout or large tracking deviations. The policy is optimized via PPO~\cite{schulman2017ppo} with a reward signal that combines tracking objectives and regularization terms. The training setup, including observation structure, reward formulation, and domain randomization, mainly follows established methods such as BeyondMimic~\cite{liao2025beyondmimic} and is detailed in the appendix.

\textbf{Data Augmentation via Motion Generation.} 
To reduce the distribution gap between motion datasets and the reference motions induced by the high-level generator, we augment the training data with generated motions. Specifically, we randomly sample text annotations from the BABEL training set to form 20-second text streams and feed them into the high-level generator, producing 5,368 motion clips totalling 31.48 hours. This approach exposes the policy to realistic variability and noise from generator outputs, enhancing its tracking performance during deployment.

\subsection{Deployment}
\label{sec:method-dep}

The system is deployed on a Unitree G1 humanoid robot~\cite{unitreeg1} with 29 degrees of freedom. The motion tracking policy executes on the robot's onboard computer at 50 Hz using ONNX Runtime~\cite{onnxruntime}, while the motion generator runs at 6.25 Hz on an external workstation equipped with an NVIDIA RTX 4090 GPU using TensorRT~\cite{tensorrt}. Communication between the two components occurs over a wired or wireless network, with a motion buffer to synchronize their execution. User textual commands are streamed to the external workstation in real time, encoded via CLIP, and incorporated into subsequent motion generation, enabling interactive control.

\begin{figure*}[t]
  \centering
  \includegraphics[width=0.9 \linewidth]{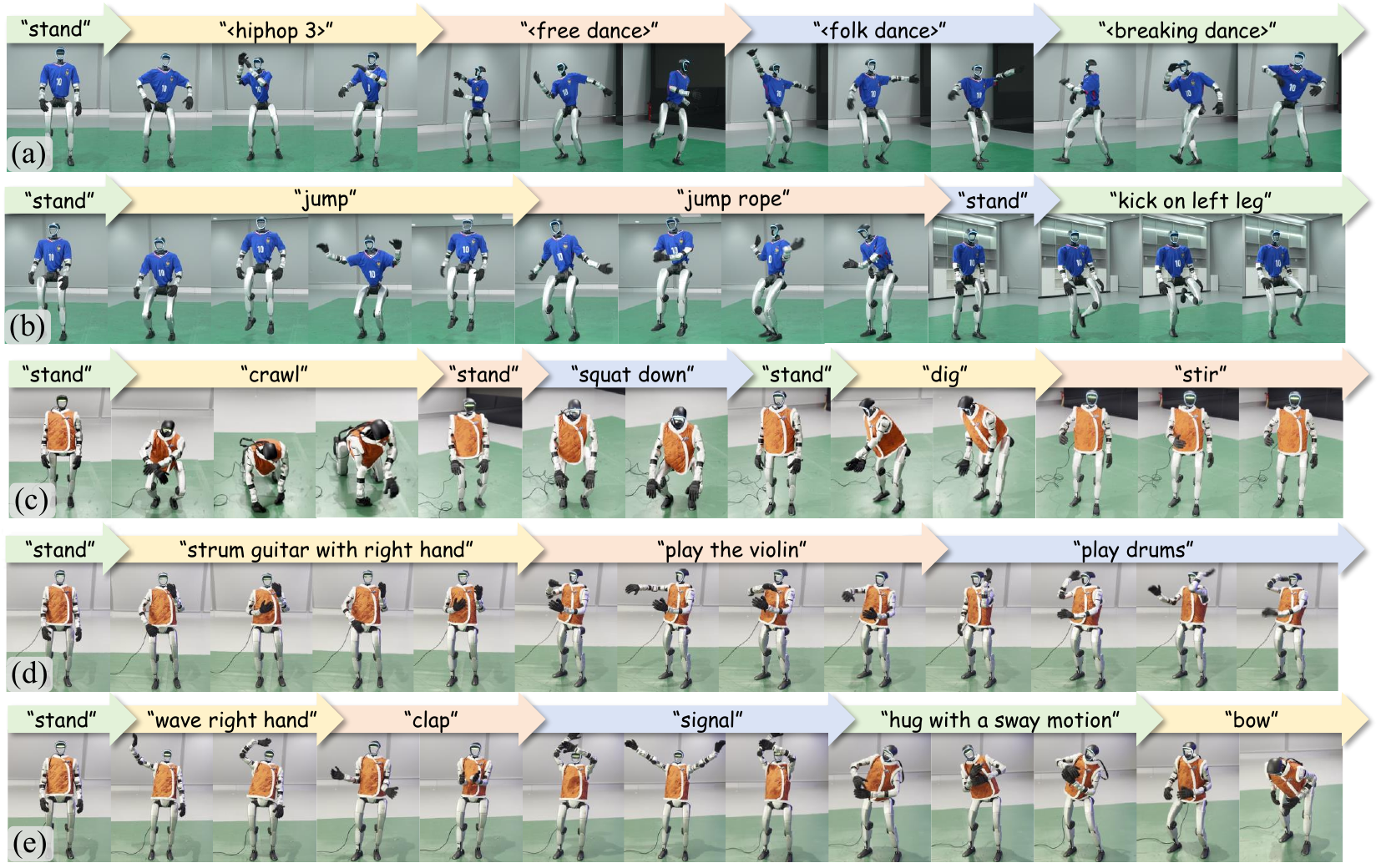}
  \caption{\textbf{Continuous diverse skill execution in the real robot.} The robot seamlessly performs a wide range of tasks, including multiple dance styles, dynamic jumping behaviors, instrument-playing motions, and expressive gestures. For complex long-horizon motions in the private dataset, the entire motion is assigned a unique label wrapped with ``$\langle \cdot \rangle$'' markers.}
  \label{fig:exp_skill_deploy}
\end{figure*}

\section{Experiments}
\label{sec:exp}

We assess the effectiveness of \ours{} through a combination of real-world deployment and large-scale simulation studies. Our experiments are designed to evaluate both the practical performance of the system and the contributions of its key components.

Specifically, the evaluation aims to answer the following research questions:

\begin{itemize}
    \item \textbf{Q1:} Can \ours{} achieve precise, stable, and responsive whole-body behaviors in the real world?
    
    \item \textbf{Q2:} Can the motion generator produce high-quality and semantically consistent motions from text commands in an interactive setting?
    
    \item \textbf{Q3:} Can the motion tracking policy robustly execute diverse reference motions, especially those produced by the generator?

\end{itemize}

\subsection{Experimental Setup}

To systematically evaluate \ours{} system, we adopt a comprehensive set of evaluation metrics covering \textbf{Motion Quality} and \textbf{Tracking Fidelity}. Formal definitions of these metrics are provided in the appendix, and different system components are evaluated using the corresponding subsets of these metrics.

\begin{itemize}
    \item \textbf{Motion Quality.}
Evaluates the overall quality of text-conditioned motion following prior work~\cite{zhao2024dartcontrol,cai2025flooddiffusion}, measured by FID and Diversity for distribution similarity to the dataset, R-precision (R@K) and multimodal distance (MM-Dist) for semantic alignment with language commands, and PJ and AUJ for temporal smoothness across command transitions.
    \item \textbf{Tracking Fidelity.} Evaluates tracking fidelity of the motion tracker under physical constraints by reporting the success rate ({Succ}) and motion discrepancy metrics between executed and reference trajectories, including global MPJPE ($E_{\text{g-mpjpe}}$, mm), root-relative MPJPE ($E_{\text{mpjpe}}$, mm), acceleration error ($E_{\text{acc}}$, mm/frame$^{2}$), and velocity error ($E_{\text{vel}}$, mm/frame), following~\cite{phc,luo2025sonic}.
\end{itemize}

For real-world deployment, the motion tracking policy is trained on a server equipped with an NVIDIA A100 GPU for approximately one week, using 150{,}000 training steps, while for simulation studies, the policy is trained for 50{,}000 steps due to computational resource constraints. The motion generation model is trained on the same hardware for approximately two days, with 200{,}000 steps for the VAE and 300{,}000 steps for the LDM. The hyperparameter setup is detailed in the appendix.

\subsection{Real-world Deployment}
\label{sec:exp-real}

To address \textbf{Q1}, we evaluate \ours{} in the real world through a combination of qualitative demonstrations, quantitative robustness tests, and real-time performance measurements.

\textbf{Complex Skill Demonstrations and Interactive Control.}
We evaluate the expressiveness and robustness of \ours{} system through a series of real-world demonstrations performed on the physical robot, as shown in Fig.~\ref{fig:teaser} and Fig.~\ref{fig:exp_skill_deploy}. The robot executes a diverse set of skills, including locomotion, dance motions, martial arts sequences, and gestures.
Beyond isolated skill execution, a distinctive feature of \ours{} is its ability to perform continuous long-horizon tasks: the robot seamlessly executes a sequence of varied commands in a single, unbroken trial. This ``one-take'' demonstration underscores the system’s capacity to maintain motion coherence and control stability.
Throughout the entire process, the interactive control capability of \ours{} allows users to modify commands on the fly, guiding the robot’s behavior and enabling smooth composition and transitions between skills.

\begin{table}[th]
\centering
\small
\caption{\textbf{Quantitative evaluation on 30-second long-horizon real-robot executions with diverse text command streams.} \ours{} achieves a strong tracking fidelity across all evaluated tasks.}
\label{tab:final_real_traj_eval}
\adjustbox{max width=1.\columnwidth, center}{
\begin{tabular}{lccccc}
\toprule
Text Command 
& Succ $\uparrow$ 
& $E_\text{g-mpjpe}$ $\downarrow$ 
& $E_\text{mpjpe}$ $\downarrow$ 
& $E_\text{acc}$ $\downarrow$ 
& $E_\text{vel}$ $\downarrow$ \\
\midrule
Random 
& 16 / 20 
& 337.169 
& 153.920 
& 1.452 
& 2.621 \\
\midrule
\multicolumn{6}{l}{\textbf{Looping}} \\
``punch'' 
& 10 / 10 
& 355.560 
& 123.598 
& 2.705 
& 4.891 \\
``wave right hand'' 
& \phantom{0}8 / 10 
& 316.480 
& \phantom{0}72.184 
& 2.757 
& 2.906 \\
``strum guitar with left hand'' 
& 10 / 10 
& 191.562 
& \phantom{0}88.662 
& 1.418 
& 1.974 \\
``play the violin'' 
& 10 / 10 
& 107.584 
& \phantom{0}33.941 
& 0.518 
& 0.750 \\
\bottomrule
\end{tabular}
}
\end{table}

\begin{figure}[ht]
  \centering
  \includegraphics[width=0.8 \linewidth]{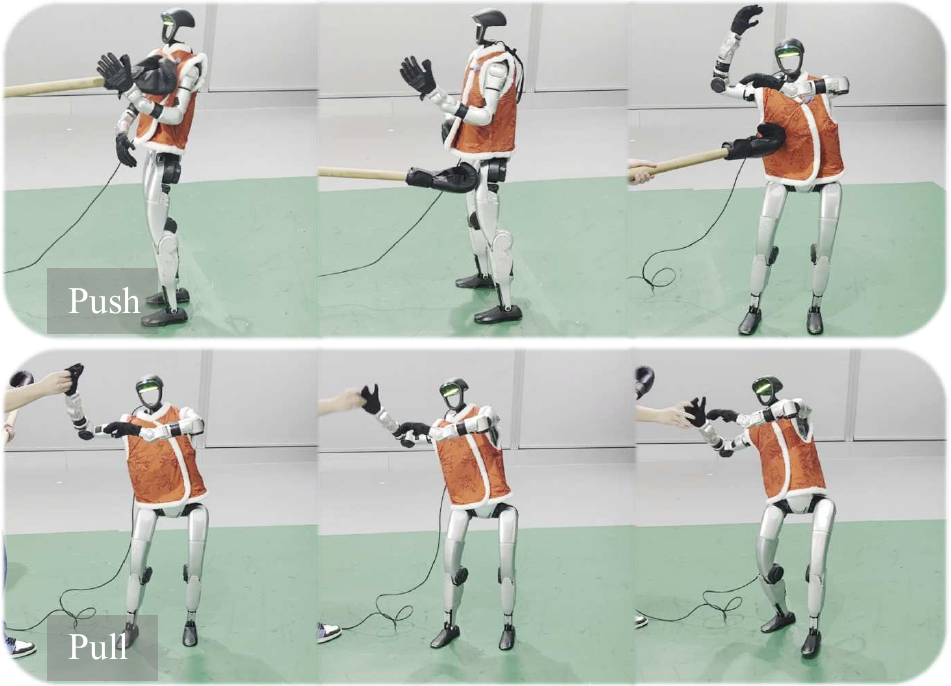}
  \caption{\textbf{Real-time recovery under external perturbations.} The robot dynamically adjusts its actions based on perturbed states to preserve stability and fulfill text-driven commands.}
  \label{fig:robustness_exp}
\end{figure}

\textbf{Robustness Analysis.}
We evaluate the system's robustness through long-horizon real-robot experiments under varying text command streams. In each 30-s trial, the robot executes either randomly sampled commands or fixed loopable instructions, such as waving or punching. As shown in Table~\ref{tab:final_real_traj_eval}, \ours{} maintains high success rates and tracking accuracy across both settings. Robustness to disturbances is further demonstrated by manually perturbing the robot during execution, where it recovers gracefully while preserving ongoing motions, as illustrated in Fig.~\ref{fig:robustness_exp}. These experiments collectively show that the system as a whole exhibits stable and reliable performance.

\textbf{Real-Time Performance.}
We evaluate the real-time performance of the system to assess its interactive capabilities. Latency is measured at key stages, from language command input to motion generation and low-level tracking. As summarized in Table~\ref{tab:latency}, the system exhibits fast and consistent response across all stages. In particular, the user interaction latency, defined as the time from typing a new command to the robot’s physical response, is $0.73$\,s in average, which is sufficiently low to support responsive real-time interaction.

\begin{table}[ht]
\centering
\caption{\textbf{Real-time performance of different stages in \ours{}.}}
\label{tab:latency}
\begin{tabular}{l S}
\toprule
\textbf{Stage} & \phantom{0000000}\textbf{Latency} \\
\midrule
Text encoding 
  & 7.64\textsubscript{$\pm$2.56}\,\text{ms} \\
Motion generator                            
  & 29.63\textsubscript{$\pm$3.56}\,\text{ms} \\
Tracking policy                             
  & 2.15\textsubscript{$\pm$0.11}\,\text{ms} \\
\midrule
User Interaction                       
  & 0.73\textsubscript{$\pm$0.10}\,\text{s} \\
\bottomrule
\end{tabular}
\end{table}

\begin{table*}[t]
    \centering
    \caption{\textbf{Comparison of motion generation with different motion representations.} Our method achieves the best overall performance across most segment- and transition-level metrics.
    Rightarrow ``$\rightarrow$'' denotes that closer alignment with the dataset reference is better.
Bold and underlined values indicate the best and second-best results, respectively, excluding the dataset.
All results are reported as mean $\pm$ standard deviation over three generator rollout seeds.}
    \adjustbox{max width=1.\textwidth, center}{
    \begin{tabular}{l|cccccc|cccc}
        \toprule
        \multirow{2}{*}{\textbf{Method}} 
        & \multicolumn{6}{c|}{\textbf{Segment}} 
        & \multicolumn{4}{c}{\textbf{Transition}} \\ 
        \cmidrule(lr){2-7} \cmidrule(lr){8-11}
        & FID $\downarrow$ & Diversity $\rightarrow$ & R@1 $\uparrow$ & R@2 $\uparrow$ & R@3 $\uparrow$ & MM-dist $\downarrow$
        & FID $\downarrow$ & Diversity $\rightarrow$ & PJ $\rightarrow$ & AUJ $\downarrow$ \\ 
        \midrule
        Dataset 
        & \phantom{0}0.000 \textsubscript{$\pm$ 0.000} & 9.150 \textsubscript{$\pm$ 0.158} & 0.357 \textsubscript{$\pm$ 0.004} & 0.556 \textsubscript{$\pm$ 0.006} 
        & 0.684 \textsubscript{$\pm$ 0.005} & 3.729 \textsubscript{$\pm$ 0.000} 
        & \phantom{0}0.000 \textsubscript{$\pm$ 0.000} & 8.138 \textsubscript{$\pm$ 0.013} & 0.005 \textsubscript{$\pm$ 0.000} & 0.000 \textsubscript{$\pm$ 0.000} \\
        \midrule
        
        DART+Retarget 
        & \phantom{0}\underline{4.837} \textsubscript{$\pm$ 0.070} & 8.012 \textsubscript{$\pm$0.082} & 0.230 \textsubscript{$\pm$0.007} & 0.372 \textsubscript{$\pm$0.020} 
        & 0.486 \textsubscript{$\pm$0.018} & 5.746 \textsubscript{$\pm$0.026} 
        & \phantom{0}\underline{5.249} \textsubscript{$\pm$0.245} & \underline{7.939} \textsubscript{$\pm$0.023} & \textbf{0.007} \textsubscript{$\pm$0.000} & \textbf{0.123} \textsubscript{$\pm$0.000} \\
        
        BeyondMimic 
        & 20.550 \textsubscript{$\pm$0.089} & 5.930 \textsubscript{$\pm$0.146} & 0.134 \textsubscript{$\pm$0.009} & 0.231 \textsubscript{$\pm$0.006} 
        & 0.320 \textsubscript{$\pm$0.013} & 7.578 \textsubscript{$\pm$0.014} 
        & 41.414 \textsubscript{$\pm$0.339} & 4.209 \textsubscript{$\pm$0.045} & 0.077 \textsubscript{$\pm$0.003} & 0.231 \textsubscript{$\pm$0.007} \\
        
        HumanML3D 
        & \phantom{0}6.599 \textsubscript{$\pm$0.031} & \underline{8.236} \textsubscript{$\pm$0.046} & 0.197 \textsubscript{$\pm$0.005} & 0.344 \textsubscript{$\pm$0.005} 
        & 0.453 \textsubscript{$\pm$0.003} & 6.292 \textsubscript{$\pm$0.023} 
        & \phantom{0}9.310 \textsubscript{$\pm$0.076} & 7.676 \textsubscript{$\pm$0.178} & 0.134 \textsubscript{$\pm$0.014} & 0.340 \textsubscript{$\pm$0.019} \\
        
        RobotMDM 
        & \phantom{0}5.134 \textsubscript{$\pm$0.131} & 8.210 \textsubscript{$\pm$0.082} & \underline{0.262} \textsubscript{$\pm$0.006} & \underline{0.427} \textsubscript{$\pm$0.004} 
        & \underline{0.556} \textsubscript{$\pm$0.010} & \underline{5.076} \textsubscript{$\pm$0.031} 
        & \phantom{0}6.514 \textsubscript{$\pm$0.178} & 7.602 \textsubscript{$\pm$0.095} & 0.032 \textsubscript{$\pm$0.001} & 0.150 \textsubscript{$\pm$0.001} \\
        \midrule
        \ours{} 
        & \phantom{0}\textbf{3.072} \textsubscript{$\pm$0.199} & \textbf{9.220} \textsubscript{$\pm$0.151} & \textbf{0.300} \textsubscript{$\pm$0.007} & \textbf{0.481} \textsubscript{$\pm$0.004} 
        & \textbf{0.607} \textsubscript{$\pm$0.008} & \textbf{4.272} \textsubscript{$\pm$0.058} 
        & \phantom{0}\textbf{3.238} \textsubscript{$\pm$0.430} & \textbf{8.226} \textsubscript{$\pm$0.079} & \underline{0.015} \textsubscript{$\pm$0.001} & \underline{0.125} \textsubscript{$\pm$0.001} \\ 
        \bottomrule
    \end{tabular}}
    \label{tab:motion_rep}
\end{table*}

\subsection{Offline Evaluation}
\label{sec:exp-sim}

To quantitatively evaluate the components of \ours{}, we conduct offline evaluations, isolating the contributions of the motion generator and the motion tracking policy. All models in these experiments are trained based on public datasets to facilitate reproducibility.

\subsubsection{Motion Generation Evaluation}
\label{sec:exp-sim-gen}

To answer \textbf{Q2}, we evaluate the motion generator independently, focusing on the effectiveness of the proposed robot-skeleton motion representation. 

\textbf{Baselines:} 
We compare the generator based on our proposed representation against several established baselines:

\begin{itemize}
    \item \textbf{DART+Retarget}: Uses the DART~\cite{zhao2024dartcontrol} model trained on the SMPL human skeleton, with the generated human motions subsequently retargeted to the robot skeleton via GMR~\cite{araujo2025gmr}.
    \item \textbf{BeyondMimic}~\cite{liao2025beyondmimic} Diffusion State: Replicates the state representation of BeyondMimic, which comprises root position, root linear velocity, root rotation, and root angular velocity relative to the initial character-yaw frame, along with local body positions and velocities relative to each character-yaw frame.
    \item \textbf{HumanML3D}-style Representation~\cite{HumanML3D}: Adapts the HumanML3D human-motion representation for the robot, including root angular velocity along the Z-axis, root linear velocity in the XY-plane, root height and rotation, body positions and velocities relative to the root, joint angles, and foot contact states. Global invariance processing is applied to enable autoregressive motion generation.
    \item \textbf{RobotMDM}~\cite{serifi2024robotmdm}: Reproduces the motion representation proposed in RobotMDM, including root height, root linear velocity in the XY-plane, root angular velocity along the Z-axis, root rotation, and joint angles and velocities. Global invariance processing is applied to facilitate autoregressive generation.
\end{itemize}

\textbf{Experimental Setup:}   
Following DART~\cite{zhao2024dartcontrol}, we evaluate text-conditioned motion generation on the BABEL validation set~\cite{babel}. Text streams are extracted from the dataset and used as conditions to generate rollout trajectories. Evaluation focuses on two aspects: (i) the complete motion segments corresponding to each text command, and (ii)  the transitions between consecutive segments.

\textbf{Results:}
Table~\ref{tab:motion_rep} summarizes the evaluation results. Our proposed motion representation achieves state-of-the-art performance both in generating motions for individual text commands and in maintaining high-quality transitions between consecutive segments.
The observed improvements over alternative representations can be attributed to two key design elements: First, the compact DoF-based motion representation reduces the inherent modeling complexity relative to HumanML3D-style representations. Second, our carefully designed incremental root-state representation improves data efficiency compared with RobotMDM and BeyondMimic.
In terms of transition smoothness, our approach is slightly below the DART+Retarget baseline, as the retargeting step introduces smoothness in joint displacements.

\begin{table*}[t]
\centering
\small
\caption{\textbf{Simulation evaluation of the motion tracker on generator-produced motion data.}
Incorporating generator-produced motions during training leads to improved performance across both tracking and semantic metrics.
Rightarrow ``$\rightarrow$'' denotes that closer alignment with the dataset reference is better.
Bold and underlined values indicate the best and second-best results, respectively, excluding the dataset.
All results are reported as mean $\pm$ standard deviation over three generator rollout seeds.}
\label{tab:final_eval_summary_gen}
\adjustbox{max width=1.\textwidth, center}{
\begin{tabular}{l|ccccc|cccccc}
\toprule
\multirow{2}{*}{\textbf{Method}}
& \multicolumn{5}{c|}{\textbf{Tracking Fidelity}}
& \multicolumn{6}{c}{\textbf{Motion Quality}} \\
\cmidrule(lr){2-6} \cmidrule(lr){7-12}
 & Succ $\uparrow$ & $E_\text{g-mpjpe}$ $\downarrow$ & $E_\text{mpjpe}$ $\downarrow$ & $E_\text{acc}$ $\downarrow$ & $E_\text{vel}$ $\downarrow$ & FID $\downarrow$ & Diversity $\rightarrow$ & R@1 $\uparrow$ & R@2 $\uparrow$ & R@3 $\uparrow$ & MM-dist $\downarrow$ \\
\midrule
Dataset
& -- & -- & -- & -- & --
& 0.000\textsubscript{$\pm$ 0.000}
& 9.150\textsubscript{$\pm$ 0.158}
& 0.357\textsubscript{$\pm$ 0.004}
& 0.556\textsubscript{$\pm$ 0.006}
& 0.684\textsubscript{$\pm$ 0.005}
& 3.729\textsubscript{$\pm$ 0.000} \\
\midrule

\ours{}-M+G (Ours) & \textbf{0.993} \textsubscript{$\pm$ 0.003} & \textbf{113.552} \textsubscript{$\pm$ 1.996} & 34.665 \textsubscript{$\pm$ 0.543} & \underline{1.906} \textsubscript{$\pm$ 0.014} & \underline{4.872} \textsubscript{$\pm$ 0.024} & \underline{3.454} \textsubscript{$\pm$ 0.225} & \underline{8.386} \textsubscript{$\pm$ 0.113} & 0.273 \textsubscript{$\pm$ 0.005} & \underline{0.453} \textsubscript{$\pm$ 0.002} & 0.574 \textsubscript{$\pm$ 0.007} & \underline{4.645} \textsubscript{$\pm$ 0.020} \\
\ours{}-G (Ours) & 0.991 \textsubscript{$\pm$ 0.002} & \underline{121.690} \textsubscript{$\pm$ 5.950} & \underline{34.570} \textsubscript{$\pm$ 0.747} & \textbf{1.857} \textsubscript{$\pm$ 0.015} & \textbf{4.786} \textsubscript{$\pm$ 0.021} & \textbf{3.371} \textsubscript{$\pm$ 0.145} & \textbf{8.409} \textsubscript{$\pm$ 0.100} & \textbf{0.281} \textsubscript{$\pm$ 0.006} & \textbf{0.464} \textsubscript{$\pm$ 0.004} & \textbf{0.577} \textsubscript{$\pm$ 0.001} & \textbf{4.642} \textsubscript{$\pm$ 0.034} \\
\ours{}-M & \underline{0.992} \textsubscript{$\pm$ 0.005} & 130.532 \textsubscript{$\pm$ 14.267} & \textbf{34.526} \textsubscript{$\pm$ 1.050} & 1.935 \textsubscript{$\pm$ 0.007} & 4.979 \textsubscript{$\pm$ 0.009} & 3.615 \textsubscript{$\pm$ 0.175} & 8.381 \textsubscript{$\pm$ 0.101} & \underline{0.281} \textsubscript{$\pm$ 0.004} & 0.452 \textsubscript{$\pm$ 0.007} & \underline{0.575} \textsubscript{$\pm$ 0.006} & 4.648 \textsubscript{$\pm$ 0.033} \\
\midrule
TWIST2 & 0.922 \textsubscript{$\pm$ 0.011} & 847.203 \textsubscript{$\pm$ 34.405} & 58.529 \textsubscript{$\pm$ 1.659} & 1.999 \textsubscript{$\pm$ 0.018} & 7.439 \textsubscript{$\pm$ 0.081} & 4.092 \textsubscript{$\pm$ 0.094} & 8.187 \textsubscript{$\pm$ 0.087} & 0.255 \textsubscript{$\pm$ 0.007} & 0.420 \textsubscript{$\pm$ 0.005} & 0.543 \textsubscript{$\pm$ 0.002} & 5.137 \textsubscript{$\pm$ 0.058} \\
GMT & 0.834 \textsubscript{$\pm$ 0.018} & 1252.235 \textsubscript{$\pm$ 35.132} & 103.138 \textsubscript{$\pm$ 1.046} & 1.934 \textsubscript{$\pm$ 0.021} & 8.776 \textsubscript{$\pm$ 0.085} & 5.142 \textsubscript{$\pm$ 0.432} & 8.010 \textsubscript{$\pm$ 0.040} & 0.240 \textsubscript{$\pm$ 0.012} & 0.397 \textsubscript{$\pm$ 0.014} & 0.503 \textsubscript{$\pm$ 0.016} & 5.441 \textsubscript{$\pm$ 0.125} \\
Any2Track & 0.714 \textsubscript{$\pm$ 0.016} & 1489.394 \textsubscript{$\pm$ 51.433} & 139.178 \textsubscript{$\pm$ 1.502} & 2.425 \textsubscript{$\pm$ 0.019} & 11.476 \textsubscript{$\pm$ 0.171} & 6.135 \textsubscript{$\pm$ 0.215} & 7.719 \textsubscript{$\pm$ 0.013} & 0.222 \textsubscript{$\pm$ 0.003} & 0.367 \textsubscript{$\pm$ 0.009} & 0.473 \textsubscript{$\pm$ 0.006} & 5.876 \textsubscript{$\pm$ 0.036} \\
\bottomrule
\end{tabular}
}
\end{table*}
\subsubsection{Motion Tracking Evaluation}

To address \textbf{Q3}, we evaluate the ability of the motion tracking policy to robustly execute diverse reference motions, with a particular focus on trajectories produced by the motion generator. This evaluation also examines the effect of augmenting the training data with generated motions on tracking robustness and generalization.

\textbf{Baselines and Ablations:} We consider the following configurations:

\begin{itemize}
    \item \textbf{\ours{}-M+G (Ours):} Tracker trained on the combined dataset of motion capture and generated motions.
    \item \textbf{\ours{}-G (Ours):} Tracker trained exclusively on generator-produced motions.
    \item \textbf{\ours{}-M:} Tracker trained exclusively on motion capture data.
    \item Pretrained general motion tracker policies from prior work: \textbf{GMT}~\cite{chen2025gmt}, \textbf{Any2Track}~\cite{zhang2025any2track}, \textbf{TWIST2}~\cite{ze2025twist2}. These models are trained on different datasets and simulation platforms; for fair comparison, we adopt the officially released pretrained checkpoints.
\end{itemize}

\textbf{Experimental Setup:} All evaluations are conducted in the MuJoCo simulator in a Sim2Sim setting. Reference trajectories are drawn from two sources: (i) \textbf{Generator-produced motion:} motions synthesized by our generator, conditioned on text streams the same as Sec.~\ref{sec:exp-sim-gen}, to approximate the deploy-time distribution; and (ii) \textbf{SnapMoGen}~\cite{hwang2025snapmogen}, a publicly available motion capture dataset, which is unseen to all baselines, to evaluate generalization. Only the test split of SnapMoGen is used for evaluation.

\begin{table}[ht]
\centering
\footnotesize
\caption{\textbf{Simulation evaluation of the motion tracker on SnapMoGen data.}
Exclusive training on generated motions shows reduced generalization, whereas training on both motion capture and generated data offers a better trade-off between deployment performance and generalization.
Bold and underlined values indicate the best and second-best results, respectively.
}

\label{tab:final_eval_summary_snap}
\adjustbox{max width=\columnwidth, center}{
\begin{tabular}{l|ccccc}
\toprule
Method & Succ $\uparrow$ & $E_\text{g-mpjpe}$ $\downarrow$ & $E_\text{mpjpe}$ $\downarrow$ & $E_\text{acc}$ $\downarrow$ & $E_\text{vel}$ $\downarrow$ \\
\midrule
\ours{}-M+G (Ours) & \underline{0.814} & \underline{394.705} & \underline{79.057} & \textbf{0.892} & \textbf{3.859} \\
\ours{}-G (Ours)   & 0.772 & 565.374 & 87.755 & \underline{0.928} & 4.310 \\
\ours{}-M          & \textbf{0.823} & \textbf{340.145} & \textbf{74.761} & 0.960 & \underline{4.020} \\
\midrule
TWIST2            & 0.634 & 1019.826 & 131.292 & 1.008 & 4.629 \\
GMT               & 0.542 & 1095.242 & 150.067 & 0.968 & 4.806 \\
Any2Track         & 0.374 & 1674.080 & 203.293 & 1.253 & 6.226 \\
\bottomrule
\end{tabular}
}
\end{table}

\textbf{Results:} 
Tables~\ref{tab:final_eval_summary_gen} and~\ref{tab:final_eval_summary_snap} summarize the motion tracking performance on generator-produced trajectories and unseen motion capture data, respectively. On generator-produced motion data, all variants of \ours{} achieve consistently high success rates and low tracking errors, indicating stable execution under text-conditioned reference motions. Among them, \ours{}-G performs best overall, while \ours{}-M+G also improves upon \ours{}-M across several tracking fidelity and motion quality metrics, suggesting that incorporating generator-produced motions into training is beneficial for aligning the tracker with the deployment-time motion distribution.

On the unseen SnapMoGen dataset, \ours{}-M achieves the strongest generalization performance, whereas \ours{}-G exhibits a noticeable performance drop, indicating limited robustness when evaluated beyond the generator’s motion distribution. The combined model, \ours{}-M+G, maintains competitive performance across both settings, balancing robustness on generator-produced trajectories and generalization to unseen motion capture data. These results reflect a trade-off between deployment-oriented robustness and generalization, motivating the use of mixed motion capture and generated data as a practical training strategy.

\section{Conclusion}

\ours{} represents a new control paradigm for humanoid robots, where natural language serves as a continuously revisable control signal rather than a one-shot task specification.
This paradigm is realized by combining a high-level motion generator with a low-level universal motion tracker, enabling smooth, whole-body motions that respond instantaneously to language commands.
Real-world experiments on the Unitree G1 robot demonstrate the system’s ability to execute behaviors ranging from simple gestures to complex skills with high fidelity, robustness, and responsiveness. Offline evaluations further indicate that carefully designed robot-skeleton motion features improve the quality of generated motions, and incorporating generator-produced motions into the training data effectively aligns the motion tracker with the generator.

Despite these strengths, \ours{} currently lacks explicit environment perception and interactive physical reasoning. Consequently, the robot cannot adapt its motions to obstacles, objects, or dynamic changes in its surroundings. Overcoming this limitation by integrating environment-aware sensing and interactive planning would enable the robot to respond flexibly to dynamic real-world scenarios. Furthermore, combining such capabilities with high-level reasoning in language models into an agent system could provide a solid foundation for autonomous, instruction-driven behavior, paving the way toward fully autonomous, general-purpose humanoid robots.

\section*{Acknowledgments}
We are grateful to Zehao Yu, Yiran Wang, Xingyi Wang, and Lei Kuang for their assistance with the real-robot experiments, and to Yiyi Cai, Yang Tang, and Yixuan Pan for their insightful discussions during the preparation of this paper.

This work was supported by the National Natural Science Foundation of China (Grant No.~62306242), the Young Elite Scientists Sponsorship Program by CAST (Grant No.~2024QNRC001), and the Yangfan Project of Shanghai (Grant No.~23YF11462200). The SJTU team was partially supported by the National Natural Science Foundation of China (Grant No.~62322603).

\bibliographystyle{plainnat}
\bibliography{references}

\clearpage
\appendices

\section{Additional Implementation Details}
\label{app:implementation}
This appendix provides supplementary implementation details that are omitted from the main text due to space constraints.

\begin{figure}[ht]
  \centering
  \includegraphics[width=0.8 \linewidth]{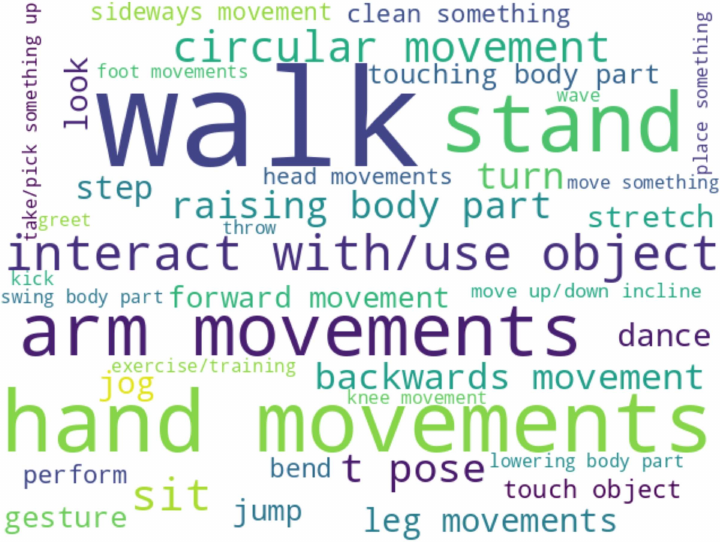}
  \caption{\textbf{Word cloud of text annotations in the BABEL dataset.}}
  \label{fig:wordcloud}
\end{figure}

\subsection{Data Preprocessing and Dataset Construction}
\label{app:impl:data}

We describe the complete data preprocessing pipeline and the construction of datasets used for training the motion generator and motion tracker. 
All datasets share a common preprocessing procedure, followed by task-specific preparation for motion generation and motion tracking.

\textbf{Shared Preprocessing Pipeline.}
\begin{enumerate}
    \item All motion sequences are first retargeted from the SMPL~\cite{loper2023smpl} human skeleton to the 29-DoF G1 humanoid robot using GMR~\cite{araujo2025gmr}. 
    \item The retargeted motions are then temporally resampled from 30~Hz to a unified frequency of 50~Hz.
    \item Following PBHC~\cite{xie2025kungfubot}, feet contact indicators are extracted from the retargeted motions using a threshold-based rule: 
    \begin{equation}
    c_t^\mathrm{left}=\mathbb{I}[\|\bm p_{t+1}^\mathrm{l\text{-}ankle}-\bm p_t^\mathrm{l\text{-}ankle}\|_2^2 <\epsilon_{\mathrm{vel}}]\cdot \mathbb{I}[p_{t,z}^\mathrm{l\text{-}ankle}<\epsilon_{\mathrm{height}}],
    \end{equation}
    where $c_t^\mathrm{left}$ represents the contact mask of the left foot and $\bm{p}_t^\mathrm{l\text{-}ankle}$ denotes the position of the left ankle joint at time $t$. $\epsilon_\mathrm{vel}=0.002$ and $\epsilon_\mathrm{height}=0.2$ are empirically chosen thresholds. Similarly for the right foot.
    \item Filtering: (1) Based on the unfiltered dataset, we train a universal motion tracking policy with privileged information, without domain randomization, and with relaxed termination conditions by removing {End-Effector Height Deviation} and {Pelvis Height Deviation}. The policy is trained for 40,000 steps, and we record the failure rate of each motion using a sliding average method. (2) We remove those motions with a tracking failure probability greater than 0.05, using the relaxed termination conditions. (3) The remaining motions are re-evaluated using standard termination conditions, and any motion that fails is removed.
    
\end{enumerate}

\textbf{Dataset Construction for Motion Generation.} For the AMASS dataset:
\begin{enumerate}
    \item 
    Motion segments are paired with textual annotations from the BABEL dataset~\cite{babel} via filename-based indexing and temporal alignment.
    The distribution of text annotations is shown in Fig.~\ref{fig:wordcloud}.

    \item 
    Mirror augmentation is applied to both motion sequences and their corresponding textual labels, effectively doubling the dataset size.
\end{enumerate}

For the proprietary dataset, which mainly contains complex long-horizon motions, we manually assign a unique label to each sequence.
To distinguish such long-form descriptions from segment-level annotations, we wrap these labels with ``$\langle \cdot \rangle$'' markers, e.g., multiple hip hop dance segments of several tens of seconds will be named as ``$\langle \text{hiphop 1} \rangle$'', ``$\langle \text{hiphop 2} \rangle$'', and so on. Since this labeling is coarse-grained, we found it allows the model to associate each label with a distinct motion style and generate stylistically consistent motions, rather than reproducing an entire reference sequence.

\textbf{Dataset Construction for Motion Tracking.}

\begin{enumerate}
    \item For data undergoing the \textbf{Shared Preprocessing Pipeline}, we further segment motions into short clips with a maximum duration of 40 seconds (i.e., 2000 frames). Specifically, each motion sequence is partitioned into multiple clips whose lengths are constrained to lie within $[100, 2000]$ frames, and adjacent clips are allowed to temporally overlap, with the overlap length constrained to lie within $[50, 200]$ frames. 
    \item Data Augmentation via Motion Generation: We sample text annotations from the BABEL training set to construct 20-second text streams. For each motion sequence, all associated text annotations are extracted, and the generation length assigned to each annotation is proportional to its temporal duration within the original motion. Using these constructed text streams, we then generate synthetic motions with our trained motion generation model.
\end{enumerate}

\subsection{Motion Generation}
\label{app:impl:generation}

\textbf{Robot Skeleton Motion Representation.}
We define a pair of deterministic forward and inverse algorithms that transform between 
a raw robot motion trajectory and a per-frame, local incremental motion representation. 
The forward algorithm maps a raw motion sequence $\{p_t, R_t, q_t, c_t\}_{t=0}^{T}$ to a feature sequence $\{f_t\}_{t=0}^{T-1}$ and an initial pose $(p_0, R_0)$. 
The inverse algorithm reconstructs the motion sequence from the feature sequence given the initial pose. 
Both transformations are invariant to global translation and yaw and are exactly invertible up to numerical precision. 
In these algorithms, $p_t \in \mathbb{R}^3$ denotes the root translation, $R_t$ the root rotation, $q_t$ the joint positions, and $c_t$ the feet contact indicators. 
The motion feature $f_t$ consists of: $\phi_t$ representing the roll and pitch angles as $(\sin \theta, \cos \theta - 1)$, $\Delta \psi_t$ the yaw increment, $\Delta p_t^{\text{local}}$ the local translation increment, $h_t$ the absolute root height, and $\Delta q_t$ the joint velocities. 
In the inverse transformation, $p'_t$ serves as an intermediate integration variable for translation, while $p_t$ represents the final reconstructed position with its $z$-component assigned by $h_t$.

\begin{algorithm}[t]
\caption{Forward Transformation: Raw Motion $\rightarrow$ Motion Features}
\label{alg:motion_forward_v3}
\begin{algorithmic}[1]
\REQUIRE Raw Motion $\{p_t, R_t, q_t, c_t\}_{t=0}^{T}$
\ENSURE Motion Features $\{f_t\}_{t=0}^{T-1}$, Initial Pose $(p_0, R_0)$
\STATE $(p_0, R_0) \leftarrow (p_0, R_0)$ 
\FOR{$t = 0$ \TO $T-1$}
    \STATE $(\mathrm{roll}_t, \mathrm{pitch}_t, \mathrm{yaw}_t) \leftarrow \textsc{QuatToEuler}(R_t)$
    \STATE $(\_, \_, \mathrm{yaw}_{t+1}) \leftarrow \textsc{QuatToEuler}(R_{t+1})$
    \STATE $s_r \leftarrow \sin(\mathrm{roll}_t), \quad c_r \leftarrow \cos(\mathrm{roll}_t) - 1$
    \STATE $s_p \leftarrow \sin(\mathrm{pitch}_t), \quad c_p \leftarrow \cos(\mathrm{pitch}_t) - 1$
    \STATE $\phi_t \leftarrow [s_r, c_r, s_p, c_p]$
    \STATE $\Delta \psi_t \leftarrow \mathrm{yaw}_{t+1} - \mathrm{yaw}_t$
    \STATE $\Delta p_t^{\text{local}} \leftarrow R_z(\mathrm{yaw}_t)^\top (p_{t+1} - p_t)$ 
    \STATE $\Delta q_t \leftarrow q_{t+1} - q_t$
    \STATE $h_t \leftarrow (p_t)_z$ 
    \STATE $f_t \leftarrow [\phi_t, \Delta \psi_t, c_t, \Delta p_t^{\text{local}}, h_t, q_t, \Delta q_t]$
\ENDFOR
\end{algorithmic}
\end{algorithm}

\begin{algorithm}[t]
\caption{Inverse Transformation: Motion Features $\rightarrow$ Raw Motion}
\label{alg:motion_inverse_v3_fixed}
\begin{algorithmic}[1]
\REQUIRE Motion features $\{f_t\}_{t=0}^{T-1}$, Initial pose $(p_{\text{init}}, R_{\text{init}})$
\ENSURE Reconstructed raw motion $\{p_t, R_t, q_t, c_t\}_{t=0}^{T-1}$
\STATE $(\_, \_, \mathrm{yaw}_0) \leftarrow \textsc{QuatToEuler}(R_{\text{init}})$
\STATE $p'_0 \leftarrow p_{\text{init}}$
\FOR{$t = 0$ \TO $T-1$}
    \STATE $[\phi_t, \Delta \psi_t, c_t, \Delta p_t^{\text{local}}, h_t, q_t, \Delta q_t] \leftarrow f_t$
    \STATE $(s_r, c_r, s_p, c_p) \leftarrow \phi_t$
    \STATE $\mathrm{roll}_t \leftarrow \arctan2(s_r, c_r + 1)$
    \STATE $\mathrm{pitch}_t \leftarrow \arctan2(s_p, c_p + 1)$
    \STATE $R_t \leftarrow \textsc{EulerToQuat}(\mathrm{roll}_t, \mathrm{pitch}_t, \mathrm{yaw}_t)$
    \STATE $p_t \leftarrow [ (p'_t)_x, (p'_t)_y, h_t ]^\top$
    \IF{$t < T-1$}
        \STATE $\mathrm{yaw}_{t+1} \leftarrow \mathrm{yaw}_t + \Delta \psi_t$
        \STATE $\Delta p_t \leftarrow R_z(\mathrm{yaw}_t) \Delta p_t^{\text{local}}$
        \STATE $p'_{t+1} \leftarrow p'_t + \Delta p_t$
    \ENDIF
\ENDFOR
\end{algorithmic}
\end{algorithm}

\textbf{Architecture.}
We design the VAE and LDM modules following the DART~\cite{zhao2024dartcontrol}. The VAE adopts a transformer-based structure, where both the encoder and decoder are implemented as stacks of transformer encoder layers augmented with skip connections. In the encoding stage, the historical and future motion sequences are concatenated, linearly projected into a high-dimensional space, and prepended with a set of learnable distribution parameters. After incorporating positional encodings, the combined sequences are processed by the transformer encoder. The output corresponding to the distribution parameters is used to parameterize a Gaussian distribution in the latent space, from which latent variables are sampled. During decoding, the sampled latent variables, together with the historical motion sequences, are used to reconstruct the future motion sequences through another transformer stack and a final linear projection layer.

The LDM is implemented as a transformer encoder that operates in the latent space. It takes four types of inputs: noisy latent variables, the diffusion timesteps, historical motion sequences, and text embeddings. Each input modality is independently embedded into a shared hidden dimension. To support classifier-free guidance, the text embeddings are randomly masked during training with a fixed probability of $0.1$. All embedded representations are concatenated into a sequence of tokens, enhanced with sinusoidal positional encodings, and fed into multiple transformer encoder layers. The output corresponding to the noisy latent token is then mapped back to the noise dimension, predicting the cleaned latent variables for the diffusion process. The detailed Motion Generator hyperparameters are shown in Table \ref{tab:generator}.

\textbf{Training Details.}

When training the VAE and LDM, we organize motion data into fixed-length motion primitives to ensure temporal continuity and stable learning.

Specifically, each motion primitive $\mathbf{P}^i$ consists of two parts:
(1) $T_{\text{history}}$ frames of historical motion, and
(2) $T_{\text{future}}$ frames of future motion to be predicted.
The historical part of $\mathbf{P}^i$ is the same as the last $T_{\text{history}}$ frames of the preceding primitive $\mathbf{P}^{i-1}$, which enforces continuity between adjacent primitives.

During data loading, we sample $N_{\text{prim}}$ consecutive primitives from a full motion sequence to form one training sample. This results in a motion clip of length $T_{\text{history}} + N_{\text{prim}} \cdot T_{\text{future}}$ frames. The model is therefore trained on an ordered sequence of primitives that together represent a coherent motion segment.

To reduce distribution mismatch between training and inference, we further adopt a self-rollout curriculum. As training progresses, with a linearly increasing probability, the history frames of a primitive are replaced by the last $T_{\text{history}}$ frames generated by the model from the previous step, instead of being directly sampled from the dataset. The replacement probability is capped at $0.8$ to maintain training stability and avoid excessive error accumulation.

For the LDM, we adopt the DDPM framework. The diffusion process gradually adds Gaussian noise to the latent motion latent $z_0$ over $K$ timesteps. The forward noising process is defined as a Markov chain:
\begin{equation}
\begin{aligned}
q(z_{1:K} | z_0) &= \prod_{k=1}^{K} q(z_k | z_{k-1}), \\
q(z_k | z_{k-1}) &= \mathcal{N}(z_k; \sqrt{1-\beta_k} z_{k-1}, \beta_k I).
\end{aligned}
\end{equation}
Here $\beta_k \in (0,1)$ is the noise schedule determined by a cosine scheduler. Let $\alpha_k = 1 - \beta_k$ and $\bar{\alpha}_k = \prod_{s=1}^{k} \alpha_s$. Then the noisy latent $z_k$ at an arbitrary timestep $k$ can be directly sampled from $z_0$ via:
\begin{equation}
\label{eq:add_noise}
z_k = \sqrt{\bar{\alpha}_k} z_0 + \sqrt{1-\bar{\alpha}_k} \epsilon, \quad \epsilon \sim \mathcal{N}(0, I).
\end{equation}
During training, we randomly sample a timestep $k \sim \mathcal{U}\{1, K\}$ and compute $z_k$ from $z_0$ using Eq.~\ref{eq:add_noise}. The LDM model $F_\theta$ then takes $z_k$, the timestep $k$, the historical motion features $f_{t-T_\text{history}:t-1}$, and the text embedding $e_t$ as inputs, and predicts the original $z_0$.

\textbf{Losses.}
The VAE is optimized with a combined loss function consisting of a reconstruction term, a lightweight KL divergence term, and a comprehensive geometric loss that ensures physical plausibility. The overall objective is defined as follows:
\begin{equation}
\mathcal{L}_{\mathrm{VAE}}=\lambda_{\text{rec}}\cdot\mathcal{L}_{\text{rec}}+\lambda_{\text{KL}}\cdot\mathcal{L}_{\mathrm{KL}}+\mathcal{L}_{\text{geo}},
\end{equation}
where $\mathcal{L}_{\mathrm{rec}}$ is the reconstruction loss. We adopt the Huber loss (smooth L1 loss) as the reconstruction objective to measure the discrepancy between the predicted future motion features $\hat{f}$ and the ground truth $f$:
\begin{equation}
    \mathcal{L}_{\mathrm{rec}}=\mathrm{Huber}(\hat{f}_{t:t+T_\text{future}-1},f_{t:t+T_\text{future}-1})
\end{equation}
This encourages the decoder to accurately reconstruct the motion sequences. To maintain a well-structured latent space without sacrificing expressiveness, we impose a KL penalty toward a standard Gaussian distribution:
\begin{equation}
\mathcal{L}_{\mathrm{KL}}=\mathrm{KL}\left(E_\phi(z\mid f_{t-T_\text{history}:t+T_\text{future}-1}))\parallel\mathcal{N}(0,I)\right)
\end{equation}
To enhance physical coherence and motion quality, we introduce a multi-term geometric loss evaluated in forward-kinematics space:
\begin{equation}
    \begin{aligned}
\mathcal{L}_{\mathrm{geo}}= & \lambda_{\text{trans}}\cdot\mathcal{L}_{\text{body\_trans}}+\lambda_{\text{rot}}\cdot\mathcal{L}_{\text{body\_rot}} \\
 & +\lambda_{\text{dof}}\cdot\mathcal{L}_{\text{dof}}+\lambda_{\text{vel}}\cdot\mathcal{L}_{\text{dof\_vel}} \\
 & +\lambda_{\text{contact}}\cdot\mathcal{L}_{\text{contact}},
\end{aligned}
\end{equation}
where $\mathcal{L}_{\text{body\_trans}}$ and $\mathcal{L}_{\text{body\_rot}}$ penalize errors in body global translation and rotation. $\mathcal{L}_{\text{dof}}$ and $\mathcal{L}_{\text{dof\_vel}}$ penalize errors in dof angles and angular velocities. $\mathcal{L}_{\mathrm{contact}}$ focuses on foot positions during contact phases to improve grounding. Each geometric term is computed using the Huber loss between predicted and ground-truth kinematic quantities, which ensemble significantly improves motion smoothness and physical realism.

The LDM is trained with the following losses:
\begin{equation}
L_{\text{LDM}}=\lambda_{\text{simple}}\cdot L_{\text{simple}}+\lambda_{\text{rec}}\cdot L_{\text{rec}}+L_{\text{geo}},
\end{equation}
where the definitions of $L_{\text{rec}}$ and $L_{\text{geo}}$ are consistent with the losses used for training the VAE and $L_{\text{simple}}$ is defined as
\begin{equation}
    \mathcal{L}_{\text{simple}}=\mathrm{Huber}(F_\theta(z_k, k, f_{t-T_\text{history}:t-1}, e_t),z_0)
\end{equation}

\textbf{Inference Details.}
During inference, we first sample a Gaussian noise $z_K \sim \mathcal{N}(0, I)$, and then iteratively denoise for timestep $k = K, K-1, \dots, 1$ to obtain the clean latent $z_0$. At each denoising step $k$, classifier-free guidance is applied to make motion latent predictions:
\begin{equation}
    \begin{aligned}
    \hat{z}_{0} & =F_\theta(z_k,k,f_{t-T_\mathrm{history}:t-1},\varnothing) \\
     & +\sigma_{\mathrm{CFG}}\cdot\left(F_\theta(z_k,k,f_{t-T_\mathrm{history}:t-1},e_t)\right. \\
     & -F_\theta(z_k,k,f_{t-T_\mathrm{history}:t-1},\varnothing)),
    \end{aligned}
\end{equation}
where $\sigma_{\text{CFG}}$ is the guidance scale. Then, the predicted $\hat{z}_0$ is used to compute the estimated noise $\epsilon_\theta$:
\begin{equation}
\epsilon_\theta = \frac{z_k - \sqrt{\bar{\alpha}_k} \hat{z}_0}{\sqrt{1-\bar{\alpha}_k}}.
\end{equation}
Finally, we sample $z_{k-1}$ from the reverse transition distribution:
\begin{equation}
z_{k-1} = \frac{1}{\sqrt{\alpha_k}} \left( z_k - \frac{1-\alpha_k}{\sqrt{1-\bar{\alpha}_k}} \epsilon_\theta \right) + \sigma_k \epsilon,
\end{equation}
where $\sigma_k^2 = \beta_k$. The final clean latent $z_0$ is then passed through the VAE decoder to produce the future motion features.



\begin{table}[h]
    \centering
    \caption{\textbf{Hyperparameters for Motion Generator.}}
    \label{tab:generator}
    \centering
    \renewcommand{\arraystretch}{1.2}
    {\fontsize{8}{8}\selectfont
    \begin{tabular}{lc}
        \toprule
        \textbf{Hyperparameter} & \textbf{Value} \\
        \midrule
        History Length $T_{\text{history}}$ & $2$ \\
        Future Length $T_{\text{future}}$ & $8$ \\
        Primitive Number $N_\text{prim}$ & $4$ \\
        Batch Size & $512$ \\
        Learning Rate & $1e-4$ \\
        Feature Dim $d_{\text{feat}}$ & $69$ \\
        $\lambda_{\text{rec}}$ & $1.0$ \\
        $\lambda_{\text{KL}}$ & $1e-4$ \\
        $\lambda_{\text{simple}}$ & $1.0$ \\
        $\lambda_{\text{trans}}$ & $0.05$ \\
        $\lambda_{\text{rot}}$ & $1e-2$ \\
        $\lambda_{\text{dof}}$ & $0.03$ \\
        $\lambda_{\text{vel}}$ & $1e-5$ \\
        $\lambda_{\text{contact}}$ & $0.01$ \\
        \midrule
        \rowcolor[HTML]{EFEFEF} \multicolumn{2}{c}{VAE Training} \\ 
        \midrule
        Hidden Dim & $512$ \\
        FFN Dim & $1024$ \\
        Latent Dim & $128$ \\
        Transformer Layers Number & $9$ \\
        Heads Number & $4$ \\
        Maximum Rollout Probability & $1.0$ \\
        \midrule
        \rowcolor[HTML]{EFEFEF} \multicolumn{2}{c}{LDM Training} \\
        \midrule
        Noise Dim & $128$ \\
        Hidden Dim & $512$ \\
        FFN Dim & $1024$ \\
        Transformer Layers Number & $8$ \\
        Heads Number & $4$ \\
        Diffusion Steps $n_{\text{denoise}}$ & $5$ \\
        Max Rollout Probability & $0.8$ \\
        \bottomrule
    \end{tabular}}
\end{table}

\subsection{Motion Tracking Policy}
\label{app:impl:tracking}

The training of the motion tracking policy adopts an asymmetric actor-critic style, where the critic network takes additional privileged information as input. The observation space, reward functions, termination conditions, and domain randomization setting are listed in Table~\ref{tab:unified_obs_space}, Table~\ref{tab:reward_table}, Table~\ref{tab:termination_table}, and Table~\ref{tab:domain_randomization}, respectively.

\begin{table*}[h]
    \centering
    \caption{\textbf{Observation space for Motion Tracker.} The \textbf{Type} column distinguishes between policy inputs (Base) and critic-only inputs (Privileged).
    The 14 key bodies include:
    Pelvis, Hips (2), Knees (2), Ankles (2), Torso, Shoulders (2), Elbows (2), and Wrists (2).
    The total dimensionality is 431 for Base observations and 557 when including Privileged inputs.}
    \label{tab:unified_obs_space}
    \begin{tabular}{l l l c}
        \toprule
        \textbf{Term} & \textbf{Description} & \textbf{Dim.} & \textbf{Type} \\
        \midrule
        Reference Motion 
        & Target joint positions and velocities over a 5-step future horizon 
        & $290 = 2 \times 29 \times 5$ & Base \\
        Reference Pelvis Position 
        & Relative pelvis positions in the base frame over a 5-step horizon 
        & $15 = 3 \times 5$ & Base \\
        Reference Pelvis Orientation
        & Pelvis orientations in the base frame using a 6D representation over a 5-step horizon 
        & $30 = 6 \times 5$ & Base \\
        Projected Gravity 
        & Gravity vector expressed in the robot base frame 
        & $3$ & Base \\
        Base Velocity 
        & Linear and angular velocities of the pelvis link. 
        & $6 = 3 + 3$ & Base \\
        Joint State 
        & Joint positions and velocities 
        & $58 = 2 \times 29$ & Base \\
        Last Action 
        & Previous joint position commands 
        & $29$ & Base \\
        \midrule
        Key Body Position 
        & Relative positions of 14 key bodies in the base frame 
        & $42 = 3 \times 14$ & Priv. \\
        Key Body Orientation 
        & Relative orientations of 14 key bodies in the base frame using a 6D representation 
        & $84 = 6 \times 14$ & Priv. \\
        \bottomrule
    \end{tabular}
\end{table*}

\begin{table*}[h]
    \centering
    \caption{\textbf{Reward functions for Motion Tracker.} Tracking terms adopt a negative exponential form reward,
    $r_i = \exp(-\|e\|^2/\sigma^2)$, where $e$ denotes the corresponding tracking error
    and $\sigma$ is the scale factor.}
    \label{tab:reward_table}
    \begin{tabular}{@{}lcl@{}}
        \toprule
        \textbf{Term} & \textbf{Weight} & \textbf{Description} \\
        \midrule
        \textit{Tracking Terms} & & \\
        Pelvis Position & $0.5$ & Global pelvis position error ($\sigma=0.3$ m) \\
        Pelvis Orientation & $0.5$ & Global pelvis orientation error ($\sigma=0.4$ rad) \\
        Key Body Position & $1.0$ & Relative body position error in the base frame ($\sigma=0.3$ m) \\
        Key Body Orientation & $1.0$ & Relative body orientation error in the base frame ($\sigma=0.4$ rad) \\
        Body Linear Velocity & $1.0$ & Global body linear velocity error in the base frame ($\sigma=1.0$ m/s) \\
        Body Angular Velocity & $1.0$ & Global body angular velocity error in the base frame ($\sigma=3.14$ rad/s) \\
        \midrule
        \textit{Regularization} & & \\
        Action Rate & $-0.1$ & L2 penalty on action differences \\
        Joint Limit & $-10.0$ & Penalty for joint position limit violations \\
        Undesired Contact & $-0.1$ & Penalty for contacts on non-end-effector links \\
        Feet Slide & $-0.3$ & Penalty on foot velocity during contact \\
        Soft Landing & $-3.0e-4$ & Penalty on impact forces for smooth footfalls \\
        Overspeed & $-1.0$ & Penalty for joint velocities exceeding 20.0 rad/s \\
        Overeffort & $-1.0$ & Penalty for joint torques exceeding actuator limits \\
        \bottomrule
    \end{tabular}
\end{table*}

\begin{table*}[h]
    \centering
    \caption{\textbf{Termination conditions for Motion Tracker.}}
    \label{tab:termination_table}
    \begin{tabular}{lp{14cm}}
        \toprule
        \textbf{Term} & \textbf{Description} \\
        \midrule
        Timeout 
        & Episode terminates when reaching the maximum duration ($T=10$ s) or when the reference motion sequence is completed. \\
        
        Pelvis Orientation Deviation 
        & Deviation in gravity-aligned orientation between the reference and robot pelvis exceeds a threshold of $0.8$, 
        measured by the difference in the projected gravity vector along the pelvis $z$-axis. \\
Pelvis Height Deviation 
& $z$-axis tracking error of the robot pelvis relative to the reference motion exceeds $0.25$ m. \\

End-Effector Height Deviation 
& $z$-axis tracking error of the robot end-effectors (ankles and wrists) relative to the reference motion exceeds $0.5$ m. \\

        \bottomrule
    \end{tabular}
\end{table*}

\begin{table*}[t]
\centering
\caption{\textbf{Domain randomization settings for Motion Tracker.}}
\label{tab:domain_randomization}
\begin{tabular}{l l}
\toprule
\textbf{Term} & \textbf{Value} \\
\midrule
Static friction 
& $\mathcal{U}(0.3, 1.6)$ \\

Dynamic friction 
& $\mathcal{U}(0.3, 1.2)$ \\

Restitution 
& $\mathcal{U}(0.0, 0.5)$ \\

Joint default position offset 
& $\mathcal{U}(-0.01, 0.01)$ rad \\

Torso CoM offset $(x, y, z)$ 
& $\mathcal{U}(-0.025, 0.025)$ m, 
  $\mathcal{U}(-0.05, 0.05)$ m, 
  $\mathcal{U}(-0.05, 0.05)$ m \\

External Push
& Interval in $\mathcal{U}(1.0, 3.0)$ s \\

\bottomrule
\end{tabular}
\end{table*}

\begin{table*}[h]
    \centering
    \caption{\textbf{Training and simulation hyperparameters for Motion Tracker.}}
    \label{tab:training_hyperparams}
    \begin{tabular}{ll}
        \toprule
        \textbf{Parameter} & \textbf{Value} \\
        \midrule
        Number of Environments & $8192$ \\
        Actor MLP Size & $[2048, 1024, 512]$ \\
        Critic MLP Size & $[2048, 1024, 512]$ \\
        Activation Function & ELU \\
        Step per Environment & $24$ \\
        PPO Init Learning Rate & $1.0e-3$ \\
        PPO Mini-batches  & $4$  \\
        PPO Learning Epochs &  $5$ \\
        PPO Clip Parameter & $0.2$ \\
        Simulation Frequency & $200$ Hz \\
        Policy Frequency & $50$ Hz \\
        \bottomrule
    \end{tabular}
\end{table*}

\section{Additional Experimental Details}
\label{app:experiment}
This appendix provides supplementary descriptions of the experimental setup, evaluation protocols, and additional results.

\subsection{Experimental Procedure}
\label{app:exp:procedure}

\textbf{Real-world Deployment.} 
In the quantitative evaluation of real-robot executions, we record the real-world robot trajectory by combining joint angles from motor sensors and the global position and orientation of the pelvis provided by the built-in odometry of the G1 robot.

In the ``Random'' command experiments, we construct a 30-second text stream by randomly sampling action descriptions from the BABEL training set. Specifically, 3–5 words are first sampled from the dataset vocabulary, and each word is assigned a random duration uniformly selected from ${6, 7, 8, 9, 10}$ seconds. To ensure a well-defined start and termination, the sequence is prefixed and suffixed with a fixed 2-second “stand” command. Based on this text stream, reference motions are generated offline using our model. Owing to the limited physical space of the experimental environment, we discard any generated motion whose maximum displacement from the initial position exceeds 2 meters. The remaining motion trajectories are then executed on the real robot.

To evaluate real-time performance, we deploy the entire system on the physical robot with motor command execution disabled and measure the latency of each module independently. The latency of the text encoding and motion generator modules is evaluated over 100 runs, whereas the tracking policy latency is measured over 1000 runs.  
To assess the latency of user interaction, we repeatedly command the real robot to switch from a ``stand'' command to a ``wave left hand'' command, and manually record the elapsed time between the moment a new text command is entered by the user and the moment the user perceives the robot initiating the corresponding motion. This latency is measured over 10 runs.

\textbf{Motion Generation Evaluation.}
We evaluate motion generation quality through a two-stage procedure consisting of motion rollout and metric-based assessment.

We first construct a evaluation set from the BABEL validation split. Each motion sequence, together with its associated textual annotations and temporal durations, is segmented into chunks with a maximum length of $200$ frames. Given these reconstructed textual prompts, the trained motion generation model is used to generate corresponding motion sequences. The generated motions are then evaluated from two complementary perspectives: segment-level quality and transition-level quality.

Segment-level quality assesses how well each generated motion segment matches its corresponding textual description, using standard motion–text consistency metrics computed independently for each segment.

Transition-level quality focuses on the temporal smoothness and coherence at textual transition boundaries. For each transition point, we extract a short motion clip consisting of $15$ frames before and after the boundary, centered at the transition. Metrics are then computed on these clips to specifically assess the quality of motion transitions.

\textbf{Motion Tracking Evaluation.}
The explanation of experimental procedure in the main text is detailed enough.

\subsection{Evaluation Metrics}
\label{app:exp:metrics}
We provide detailed definitions of the evaluation metrics used in our experiments.

\textbf{Motion Quality.} 
We train the motion and text feature extractors based on TMR~\cite{petrovich2023tmr} to evaluate the quality of the generated motions. Specifically, we construct the BABEL dataset into individual text-motion pairs, with a maximum motion length of 200 frames. Through contrastive learning, we train motion and text feature extractors to produce closely aligned embeddings for semantically matching text-motion pairs.

\begin{itemize}
    \item \textbf{Frechet Inception Distance (FID).} FID is the principal metric for assessing the distributional similarity between generated and real motions in the extracted feature space. Let $P_g$ represent the empirical distribution of generated motion features and $P_r$ the empirical distribution of real motion features, with $\mu_g$, $\Sigma_g$ and $\mu_r$, $\Sigma_r$ as their respective means and covariances. The FID is computed as:
    \begin{equation}
        \mathrm{FID}=\|\mu_g-\mu_r\|^2+\mathrm{Tr}\left(\Sigma_g+\Sigma_r-2(\Sigma_g\Sigma_r)^{1/2}\right)
    \end{equation}
    
    \item \textbf{Diversity (DIV).} Diversity measures the variance across the entire set of generated motions. To compute DIV, all generated motions are randomly split into two subsets of equal size $D$, with feature vectors $\{x_i\}_{i=1}^D$ and $\{x_i^{\prime}\}_{i=1}^D$. The metric is defined as:
    \begin{equation}
        \text{Diversity} =\frac{1}{X_d}\sum_{i=1}^{X_d}\|x_i-x_i^{\prime}\|
    \end{equation}
    
    \item \textbf{R-precision (R@K).} R@K evaluates text-to-motion semantic alignment by measuring retrieval accuracy within the feature space. Each time, a batch of $M=32$ motion-text pairs is sampled from the generated motion.  
    For each motion, its corresponding text is used as a query to retrieve the most similar motions from the batch, by ranking motions based on Euclidean distance in the extracted feature space learned by the text and motion extractors. R@K is then defined as the percentage of queries for which the correct motion appears among the top-K retrieved results. 
    
    \item \textbf{Multimodal distance (MM-Dist).} MM-Dist directly measures the average proximity between a generated motion and its corresponding text in the feature space. For a batch of $M=32$ generated motions with their corresponding text descriptions, let $\{h_i\}_{i=1}^M$ and $\{x_i\}_{i=1}^M$ be their respective text and motion feature vectors. MM-Dist is computed as:
    \begin{equation}
        \text{MM-Dist}=\frac{1}{M}\sum_{i=1}^M\|x_i-h_i\|
    \end{equation}

    \item \textbf{Peak Jerk (PJ).} PJ captures the most extreme instantaneous jerk occurring during a motion transition. It is defined as the maximum absolute jerk value observed across all joints $K$ and all time steps within the transition sequence of length $L_\text{tr}$:
    \begin{equation}
        \text{PJ}=\max_{\begin{array}{c}1\leq i\leq K \\1\leq t\leq L_{tr}\end{array}}|j_i(t)|_1,
    \end{equation}
    where $j_i(t)$ is the jerk of joint $i$ at time $t$.

    \item \textbf{Area Under the Jerk (AUJ).} AUJ measures the cumulative deviation from natural smoothness over the entire transition. It aggregates the absolute difference between the instantaneous jerk at each frame and the dataset's jerk level $j_{avg}$. The metric is computed as:
    \begin{equation}
        \text{AUJ}=\sum_{t=1}^{L_{tr}}\max_{1\leq i\leq K}|j_{i}(t)-j_{avg}|_{1}
    \end{equation}
\end{itemize}

\textbf{Tracking Fidelity.}
\begin{itemize}
    \item 

\textbf{Success Rate (Succ).}
In the real-world experiments, a trial is considered successful if the robot completes the full motion execution without falling, encountering hardware malfunctions, or exceeding the predefined safety boundaries of the workspace.

In the simulation experiments, success rate measures the proportion of motion sequences whose root-relative tracking error remains below a predefined threshold over the entire trajectory. For each sequence $i$, we define the maximum root-relative mean per-link position error as
\begin{equation}
e_i
= \max_{t}
\frac{1}{J}
\sum_{j=1}^{J}
\|(\mathbf{p}_t^{\text{pol}}[j] - \mathbf{p}_t^{\text{pol}}[0])
 - (\mathbf{p}_t^{\text{ref}}[j] - \mathbf{p}_t^{\text{ref}}[0])\|_2 .
\end{equation}
The success rate is then computed as
\begin{equation}
\text{Succ}
= \frac{1}{N} \sum_{i=1}^{N} \mathbb{I}(e_i \le \theta),
\end{equation}
where $N$ is the number of test sequences and $\theta = 0.3$\,m.

\item
\textbf{Global MPJPE ($E_{\text{g-mpjpe}}$)}. Global Mean Per Joint Position Error measures the average 3D position error across all links:

\begin{equation}
E_{\text{g-mpjpe}} = \frac{1000}{T \cdot J} \sum_{t=1}^{T} \sum_{j=1}^{J} \|\mathbf{p}_t^{\text{pol}}[j] - \mathbf{p}_t^{\text{ref}}[j]\|_2
\end{equation}

where $T$ is the number of time steps, $J$ is the total number of links, and the factor of 1000 converts from meters to millimeters.

Although termed \emph{Per Joint}, the name follows conventions established in prior virtual character animation literature; in our robot setting, the metric is computed over all corresponding links.

\item
\textbf{Root-relative MPJPE ($E_{\text{mpjpe}}$)}. Root-relative MPJPE measures position errors relative to the root link, eliminating systematic translation errors:

\begin{equation}
E_{\text{mpjpe}} = \frac{1000}{T \cdot J} \sum_{t=1}^{T} \sum_{j=1}^{J} \|(\mathbf{p}_t^{\text{pol}}[j] - \mathbf{p}_t^{\text{pol}}[0]) - (\mathbf{p}_t^{\text{ref}}[j] - \mathbf{p}_t^{\text{ref}}[0])\|_2
\end{equation}

where joint 0 represents the root (pelvis) link.

\item
\textbf{Velocity Error ($E_{\text{vel}}$)}. Velocity error measures temporal consistency of motion execution:

\begin{equation}
E_{\text{vel}} = \frac{1000}{T-1} \sum_{t=1}^{T-1} \|\mathbf{v}_t^{\text{pol}} - \mathbf{v}_t^{\text{ref}}\|_2
\end{equation}

where velocities are computed as $\mathbf{v}_t = {\mathbf{p}_{t+1} - \mathbf{p}_t}$.

\item
\textbf{Acceleration Error ($E_{\text{acc}}$)}. Acceleration error captures higher-order temporal dynamics:

\begin{equation}
E_{\text{acc}} = \frac{1000}{T-2} \sum_{t=1}^{T-2} \|\mathbf{a}_t^{\text{pol}} - \mathbf{a}_t^{\text{ref}}\|_2
\end{equation}

where accelerations are computed as $\mathbf{a}_t = {\mathbf{v}_{t+1} - \mathbf{v}_t}$.

\end{itemize}

\begin{table*}[t]
    \centering
    \caption{\textbf{Comparison of motion representation components used by different methods.} \greencheck\ indicates the representation is used, \redcross\ indicates it is not used. Components are reported based on the actual feature vectors used by each method, regardless of semantic overlap between each other. For the following tables, in the Unitree G1 skeleton, the term ``Body'' refers to 29 body links, excluding the root link, whereas in the SMPL skeleton it corresponds to 22 spherical joints. The ``Base Frame'' denotes the coordinate frame attached to the root link. In contrast, the ``Character Frame'' is a local frame aligned with the root’s yaw: its origin coincides with the root position, its yaw matches the root yaw, and its roll and pitch are set to zero. }
    \adjustbox{max width=1.\textwidth, center}{
    \begin{tabular}{l|l|cccc|c}
        \toprule
        \textbf{Components} & 
        \textbf{Dimension} &
        \textbf{DART+Retarget} & \textbf{BeyondMimic} & \textbf{HumanML3D} & \textbf{RobotMDM} & \textbf{\ours{}} \\
        \midrule
        Is Robot-Skeleton Generation & - & \redcross & \greencheck & \greencheck & \greencheck & \greencheck \\
        \midrule
        Root Trigonometric Encoding of Roll and Pitch & 4 & \redcross & \redcross & \redcross & \redcross & \greencheck \\

        Root Rotation 6D Representation & 6 & \greencheck & \redcross & \redcross & \redcross & \redcross \\

        Root Rotation Quaternion Representation & 4 & \redcross & \greencheck & \greencheck & \greencheck & \redcross \\

        Difference of Root Rotation 6D Representation & 6 & \greencheck & \redcross & \redcross & \redcross & \redcross \\

        Root Yaw Velocity & 1 & \redcross & \redcross & \redcross & \redcross & \greencheck \\
        Root Angular Velocity in the Z-axis & 1 & \redcross & \redcross & \greencheck & \greencheck & \redcross \\
        Root Angular Velocity & 3 & \redcross & \greencheck & \redcross & \redcross & \redcross \\ 
        
        \midrule
        
        Root Position & 3 & \greencheck & \greencheck & \redcross & \redcross & \redcross \\
        Root Height & 1 & \redcross & \redcross & \greencheck & \greencheck & \greencheck \\
        Root Linear Velocity in the Character Frame & 3 & \greencheck & \greencheck & \redcross & \redcross & \greencheck \\
        Root Linear Velocity in the XY-plane in the Character Frame & 2 & \redcross & \redcross & \greencheck & \greencheck & \redcross \\      

        \midrule
        
        Dof Angle & 29 & \redcross & \redcross & \greencheck & \greencheck & \greencheck \\
        Dof Velocity & 29 & \redcross & \redcross & \redcross & \greencheck & \greencheck \\
        Body Rotation 6D Representation in the Base Frame & 126 for SMPL & \greencheck & \redcross & \redcross & \redcross & \redcross \\        
        Body Position in the Character Frame & 87 (66 for SMPL) & \greencheck & \greencheck & \greencheck & \redcross & \redcross \\
        Body Linear Velocity in the Character Frame & 87 (66 for SMPL) & \greencheck & \greencheck & \greencheck & \redcross & \redcross \\
        Foot Contact & 2 & \redcross & \redcross & \greencheck & \redcross & \greencheck \\
        \midrule
        Total Dim & - & 276 & 187 & 213 & 66 & 69 \\
        \bottomrule
    \end{tabular}}
    \label{tab:motion_representation_components}
\end{table*}

\subsection{Baseline Methods}
\label{app:exp:baselines}
This subsection describes the baseline methods used for comparison.

\textbf{Motion Representation.}
Table~\ref{tab:motion_representation_components} illustrates the differences in motion feature representations between \ours{} and the baselines. 

\textbf{Motion Tracking.}
\begin{itemize}
    \item 

TWIST2~\cite{ze2025twist2} is a scalable and holistic humanoid data collection system designed to generate high-quality, whole-body demonstrations. Within its hierarchical framework, the low-level controller is implemented as a task-agnostic motion tracking engine. We adopt the officially released checkpoint of the low-level controller to conduct the experiment.

    \item 
GMT~\cite{chen2025gmt} trains a single unified policy for general humanoid motion tracking through two core components: an adaptive sampling strategy that balances easy and difficult motions during training, and a Motion Mixture-of-Experts architecture that specializes across different regions of the motion manifold.

    \item 
Any2Track~\cite{zhang2025any2track} employs a two-stage reinforcement learning framework that reformulates dynamics adaptability as an additional capability on top of basic action execution. The system consists of AnyTracker, a general motion tracker for handling diverse, highly dynamic motions within a single policy, and AnyAdapter, a history-informed adaptation module that provides online dynamics adaptability. We adopt the officially released checkpoint of AnyTracker, the model trained in the first stage to conduct the experiment, since the second stage is not open-source.

\end{itemize}

\begin{table*}[t]
    \centering
    \caption{\textbf{Ablation study on key hyperparameters for Motion Generation.} All results are reported as mean $\pm$ standard deviation over three generator rollout seeds.}
    \adjustbox{max width=1.\textwidth, center}{
    \begin{tabular}{l|cccccc|cccc}
        \toprule
        \multirow{2}{*}{\textbf{Setting}} 
        & \multicolumn{6}{c|}{\textbf{Segment}} 
        & \multicolumn{4}{c}{\textbf{Transition}} \\ 
        \cmidrule(lr){2-7} \cmidrule(lr){8-11}
        & FID $\downarrow$ & Diversity $\rightarrow$ & R@1 $\uparrow$ & R@2 $\uparrow$ & R@3 $\uparrow$ 
        & MM-dist $\downarrow$ & FID $\downarrow$ & Diversity $\rightarrow$ & PJ $\rightarrow$ & AUJ $\downarrow$ \\ 
        \midrule

        Dataset 
        & 0.000 \textsubscript{$\pm$ 0.000} & 9.150 \textsubscript{$\pm$ 0.158} & 0.357 \textsubscript{$\pm$ 0.004} & 0.556 \textsubscript{$\pm$ 0.006} 
        & 0.684 \textsubscript{$\pm$ 0.005} & 3.729 \textsubscript{$\pm$ 0.000} 
        & 0.000 \textsubscript{$\pm$ 0.000} & 8.138 \textsubscript{$\pm$ 0.013} & 0.005 \textsubscript{$\pm$ 0.000} & 0.000 \textsubscript{$\pm$ 0.000} \\
        \midrule
        
        $T_{\text{history}}=1$ 
        & 2.972 \textsubscript{$\pm$ 0.032} & 9.082 \textsubscript{$\pm$ 0.139} & 0.283 \textsubscript{$\pm$ 0.011} & 0.461 \textsubscript{$\pm$ 0.002} & 0.588 \textsubscript{$\pm$ 0.004} 
        & 4.321 \textsubscript{$\pm$ 0.107} & 4.107 \textsubscript{$\pm$ 0.268} & 8.528 \textsubscript{$\pm$ 0.139} & 0.017 \textsubscript{$\pm$ 0.002} & 0.128 \textsubscript{$\pm$ 0.002} \\

        $T_{\text{history}}=2$ (Ours) 
        & 3.072 \textsubscript{$\pm$ 0.199} & 9.220 \textsubscript{$\pm$ 0.151} & 0.300 \textsubscript{$\pm$ 0.007} & 0.481 \textsubscript{$\pm$ 0.004} & 0.607 \textsubscript{$\pm$ 0.008} 
        & 4.272 \textsubscript{$\pm$ 0.058} & 3.238 \textsubscript{$\pm$ 0.430} & 8.226 \textsubscript{$\pm$ 0.079} & 0.015 \textsubscript{$\pm$ 0.001} & 0.125 \textsubscript{$\pm$ 0.001} \\
        
        $T_{\text{history}}=3$ 
        & 3.718 \textsubscript{$\pm$ 0.091} & 9.218 \textsubscript{$\pm$ 0.094} & 0.287 \textsubscript{$\pm$ 0.012} & 0.467 \textsubscript{$\pm$ 0.009} & 0.594 \textsubscript{$\pm$ 0.007} 
        & 4.366 \textsubscript{$\pm$ 0.062} & 3.695 \textsubscript{$\pm$ 0.167} & 8.470 \textsubscript{$\pm$ 0.085} & 0.016 \textsubscript{$\pm$ 0.001} & 0.126 \textsubscript{$\pm$ 0.001} \\
        
        $T_{\text{history}}=5$ 
        & 4.713 \textsubscript{$\pm$ 0.046} & 9.060 \textsubscript{$\pm$ 0.079} & 0.279 \textsubscript{$\pm$ 0.012} & 0.453 \textsubscript{$\pm$ 0.001} & 0.566 \textsubscript{$\pm$ 0.006} 
        & 4.508 \textsubscript{$\pm$ 0.049} & 4.750 \textsubscript{$\pm$ 0.121} & 8.432 \textsubscript{$\pm$ 0.154} & 0.019 \textsubscript{$\pm$ 0.000} & 0.127 \textsubscript{$\pm$ 0.001} \\

        \midrule
        
        $T_{\text{future}}=4$ 
        & 3.339 \textsubscript{$\pm$ 0.089} & 8.808 \textsubscript{$\pm$ 0.104} & 0.256 \textsubscript{$\pm$ 0.010} & 0.431 \textsubscript{$\pm$ 0.011} & 0.549 \textsubscript{$\pm$ 0.003} 
        & 4.747 \textsubscript{$\pm$ 0.023} & 2.904 \textsubscript{$\pm$ 0.140} & 8.301 \textsubscript{$\pm$ 0.136} & 0.013 \textsubscript{$\pm$ 0.000} & 0.127 \textsubscript{$\pm$ 0.000} \\
        
        $T_{\text{future}}=6$ 
        & 2.930 \textsubscript{$\pm$ 0.098} & 9.096 \textsubscript{$\pm$ 0.027} & 0.283 \textsubscript{$\pm$ 0.010} & 0.467 \textsubscript{$\pm$ 0.011} & 0.584 \textsubscript{$\pm$ 0.006} 
        & 4.481 \textsubscript{$\pm$ 0.027} & 2.910 \textsubscript{$\pm$ 0.193} & 8.453 \textsubscript{$\pm$ 0.135} & 0.016 \textsubscript{$\pm$ 0.001} & 0.126 \textsubscript{$\pm$ 0.001} \\

        $T_{\text{future}}=8$ (Ours) 
        & 3.072 \textsubscript{$\pm$ 0.199} & 9.220 \textsubscript{$\pm$ 0.151} & 0.300 \textsubscript{$\pm$ 0.007} & 0.481 \textsubscript{$\pm$ 0.004} & 0.607 \textsubscript{$\pm$ 0.008} 
        & 4.272 \textsubscript{$\pm$ 0.058} & 3.238 \textsubscript{$\pm$ 0.430} & 8.226 \textsubscript{$\pm$ 0.079} & 0.015 \textsubscript{$\pm$ 0.001} & 0.125 \textsubscript{$\pm$ 0.001} \\
        
        $T_{\text{future}}=10$ 
        & 4.126 \textsubscript{$\pm$ 0.073} & 9.206 \textsubscript{$\pm$ 0.070} & 0.326 \textsubscript{$\pm$ 0.004} & 0.501 \textsubscript{$\pm$ 0.004} & 0.624 \textsubscript{$\pm$ 0.004} 
        & 3.854 \textsubscript{$\pm$ 0.015} & 5.467 \textsubscript{$\pm$ 0.192} & 8.606 \textsubscript{$\pm$ 0.163} & 0.028 \textsubscript{$\pm$ 0.002} & 0.186 \textsubscript{$\pm$ 0.002} \\
        
        $T_{\text{future}}=12$ 
        & 3.161 \textsubscript{$\pm$ 0.062} & 9.213 \textsubscript{$\pm$ 0.175} & 0.300 \textsubscript{$\pm$ 0.003} & 0.478 \textsubscript{$\pm$ 0.013} & 0.603 \textsubscript{$\pm$ 0.010} 
        & 4.101 \textsubscript{$\pm$ 0.025} & 4.080 \textsubscript{$\pm$ 0.244} & 8.496 \textsubscript{$\pm$ 0.125} & 0.016 \textsubscript{$\pm$ 0.001} & 0.127 \textsubscript{$\pm$ 0.002} \\
        \midrule
        
        $N_{\text{prim}}=1$ 
        & 3.044 \textsubscript{$\pm$ 0.204} & 8.391 \textsubscript{$\pm$ 0.066} & 0.196 \textsubscript{$\pm$ 0.005} & 0.346 \textsubscript{$\pm$ 0.011} & 0.452 \textsubscript{$\pm$ 0.014} 
        & 6.245 \textsubscript{$\pm$ 0.058} & 3.074 \textsubscript{$\pm$ 0.183} & 8.047 \textsubscript{$\pm$ 0.275} & 0.013 \textsubscript{$\pm$ 0.001} & 0.123 \textsubscript{$\pm$ 0.001} \\
        
        $N_{\text{prim}}=2$ 
        & 2.578 \textsubscript{$\pm$ 0.037} & 8.861 \textsubscript{$\pm$ 0.117} & 0.256 \textsubscript{$\pm$ 0.011} & 0.419 \textsubscript{$\pm$ 0.011} & 0.547 \textsubscript{$\pm$ 0.015} 
        & 4.990 \textsubscript{$\pm$ 0.038} & 2.277 \textsubscript{$\pm$ 0.197} & 8.265 \textsubscript{$\pm$ 0.087} & 0.011 \textsubscript{$\pm$ 0.000} & 0.122 \textsubscript{$\pm$ 0.000} \\

        $N_{\text{prim}}=4$ (Ours)
        & 3.072 \textsubscript{$\pm$ 0.199} & 9.220 \textsubscript{$\pm$ 0.151} & 0.300 \textsubscript{$\pm$ 0.007} & 0.481 \textsubscript{$\pm$ 0.004} & 0.607 \textsubscript{$\pm$ 0.008} 
        & 4.272 \textsubscript{$\pm$ 0.058} & 3.238 \textsubscript{$\pm$ 0.430} & 8.226 \textsubscript{$\pm$ 0.079} & 0.015 \textsubscript{$\pm$ 0.001} & 0.125 \textsubscript{$\pm$ 0.001} \\
        
        $N_{\text{prim}}=6$ 
        & 3.733 \textsubscript{$\pm$ 0.088} & 9.141 \textsubscript{$\pm$ 0.092} & 0.307 \textsubscript{$\pm$ 0.008} & 0.492 \textsubscript{$\pm$ 0.010} & 0.616 \textsubscript{$\pm$ 0.012} 
        & 4.011 \textsubscript{$\pm$ 0.012} & 4.802 \textsubscript{$\pm$ 0.205} & 8.421 \textsubscript{$\pm$ 0.054} & 0.022 \textsubscript{$\pm$ 0.002} & 0.133 \textsubscript{$\pm$ 0.002} \\
        
        $N_{\text{prim}}=10$ 
        & 3.344 \textsubscript{$\pm$ 0.098} & 9.111 \textsubscript{$\pm$ 0.102} & 0.314 \textsubscript{$\pm$ 0.010} & 0.482 \textsubscript{$\pm$ 0.014} & 0.611 \textsubscript{$\pm$ 0.005} 
        & 3.897 \textsubscript{$\pm$ 0.060} & 5.131 \textsubscript{$\pm$ 0.272} & 8.315 \textsubscript{$\pm$ 0.064} & 0.023 \textsubscript{$\pm$ 0.001} & 0.134 \textsubscript{$\pm$ 0.001} \\
        \midrule
        
        $\text{Hidden}=128$ 
        & 4.030 \textsubscript{$\pm$ 0.242} & 9.493 \textsubscript{$\pm$ 0.126} & 0.255 \textsubscript{$\pm$ 0.006} & 0.435 \textsubscript{$\pm$ 0.005} & 0.558 \textsubscript{$\pm$ 0.006} 
        & 4.640 \textsubscript{$\pm$ 0.032} & 5.353 \textsubscript{$\pm$ 0.208} & 8.903 \textsubscript{$\pm$ 0.126} & 0.015 \textsubscript{$\pm$ 0.001} & 0.126 \textsubscript{$\pm$ 0.002} \\
        
        $\text{Hidden}=256$ 
        & 3.110 \textsubscript{$\pm$ 0.082} & 9.089 \textsubscript{$\pm$ 0.123} & 0.273 \textsubscript{$\pm$ 0.009} & 0.448 \textsubscript{$\pm$ 0.004} & 0.582 \textsubscript{$\pm$ 0.009} 
        & 4.487 \textsubscript{$\pm$ 0.051} & 2.506 \textsubscript{$\pm$ 0.197} & 8.279 \textsubscript{$\pm$ 0.159} & 0.011 \textsubscript{$\pm$ 0.000} & 0.123 \textsubscript{$\pm$ 0.000} \\

        $\text{Hidden}=512$ (Ours) 
        & 3.072 \textsubscript{$\pm$ 0.199} & 9.220 \textsubscript{$\pm$ 0.151} & 0.300 \textsubscript{$\pm$ 0.007} & 0.481 \textsubscript{$\pm$ 0.004} & 0.607 \textsubscript{$\pm$ 0.008} 
        & 4.272 \textsubscript{$\pm$ 0.058} & 3.238 \textsubscript{$\pm$ 0.430} & 8.226 \textsubscript{$\pm$ 0.079} & 0.015 \textsubscript{$\pm$ 0.001} & 0.125 \textsubscript{$\pm$ 0.001} \\
        
        $\text{Hidden}=1024$ 
        & 3.552 \textsubscript{$\pm$ 0.194} & 9.028 \textsubscript{$\pm$ 0.167} & 0.277 \textsubscript{$\pm$ 0.009} & 0.464 \textsubscript{$\pm$ 0.012} & 0.598 \textsubscript{$\pm$ 0.007} 
        & 4.324 \textsubscript{$\pm$ 0.040} & 3.607 \textsubscript{$\pm$ 0.076} & 8.451 \textsubscript{$\pm$ 0.100} & 0.014 \textsubscript{$\pm$ 0.000} & 0.124 \textsubscript{$\pm$ 0.001} \\
        
        $\text{Hidden}=2048$ 
        & 3.688 \textsubscript{$\pm$ 0.131} & 9.154 \textsubscript{$\pm$ 0.211} & 0.285 \textsubscript{$\pm$ 0.010} & 0.449 \textsubscript{$\pm$ 0.009} & 0.581 \textsubscript{$\pm$ 0.004} 
        & 4.509 \textsubscript{$\pm$ 0.038} & 4.326 \textsubscript{$\pm$ 0.385} & 8.633 \textsubscript{$\pm$ 0.161} & 0.013 \textsubscript{$\pm$ 0.000} & 0.125 \textsubscript{$\pm$ 0.000} \\
        \midrule
        
        $\text{Layers}=4$ 
        & 4.186 \textsubscript{$\pm$ 0.098} & 9.315 \textsubscript{$\pm$ 0.089} & 0.303 \textsubscript{$\pm$ 0.010} & 0.477 \textsubscript{$\pm$ 0.006} & 0.597 \textsubscript{$\pm$ 0.004} 
        & 4.225 \textsubscript{$\pm$ 0.024} & 5.151 \textsubscript{$\pm$ 0.280} & 8.584 \textsubscript{$\pm$ 0.117} & 0.014 \textsubscript{$\pm$ 0.000} & 0.124 \textsubscript{$\pm$ 0.000} \\
        
        $\text{Layers}=6$ 
        & 3.868 \textsubscript{$\pm$ 0.203} & 9.213 \textsubscript{$\pm$ 0.077} & 0.291 \textsubscript{$\pm$ 0.005} & 0.462 \textsubscript{$\pm$ 0.009} & 0.596 \textsubscript{$\pm$ 0.004} 
        & 4.275 \textsubscript{$\pm$ 0.034} & 4.609 \textsubscript{$\pm$ 0.365} & 8.701 \textsubscript{$\pm$ 0.187} & 0.015 \textsubscript{$\pm$ 0.001} & 0.127 \textsubscript{$\pm$ 0.004} \\

        $\text{Layers}=8$ (Ours) 
        & 3.072 \textsubscript{$\pm$ 0.199} & 9.220 \textsubscript{$\pm$ 0.151} & 0.300 \textsubscript{$\pm$ 0.007} & 0.481 \textsubscript{$\pm$ 0.004} & 0.607 \textsubscript{$\pm$ 0.008} 
        & 4.272 \textsubscript{$\pm$ 0.058} & 3.238 \textsubscript{$\pm$ 0.430} & 8.226 \textsubscript{$\pm$ 0.079} & 0.015 \textsubscript{$\pm$ 0.001} & 0.125 \textsubscript{$\pm$ 0.001} \\
        
        $\text{Layers}=10$ 
        & 3.342 \textsubscript{$\pm$ 0.082} & 9.030 \textsubscript{$\pm$ 0.051} & 0.296 \textsubscript{$\pm$ 0.003} & 0.476 \textsubscript{$\pm$ 0.002} & 0.601 \textsubscript{$\pm$ 0.007} 
        & 4.254 \textsubscript{$\pm$ 0.021} & 3.272 \textsubscript{$\pm$ 0.199} & 8.205 \textsubscript{$\pm$ 0.056} & 0.012 \textsubscript{$\pm$ 0.000} & 0.123 \textsubscript{$\pm$ 0.001} \\
        \midrule
        
        $\sigma_\text{CFG}=3$ 
        & 2.537 \textsubscript{$\pm$ 0.056} & 8.919 \textsubscript{$\pm$ 0.227} & 0.259 \textsubscript{$\pm$ 0.003} & 0.441 \textsubscript{$\pm$ 0.004} & 0.567 \textsubscript{$\pm$ 0.006} 
        & 4.667 \textsubscript{$\pm$ 0.048} & 1.493 \textsubscript{$\pm$ 0.168} & 8.036 \textsubscript{$\pm$ 0.092} & 0.013 \textsubscript{$\pm$ 0.001} & 0.124 \textsubscript{$\pm$ 0.001} \\
        
        $\sigma_\text{CFG}=4$ 
        & 2.818 \textsubscript{$\pm$ 0.118} & 9.058 \textsubscript{$\pm$ 0.133} & 0.295 \textsubscript{$\pm$ 0.008} & 0.480 \textsubscript{$\pm$ 0.007} & 0.597 \textsubscript{$\pm$ 0.005} 
        & 4.384 \textsubscript{$\pm$ 0.052} & 2.499 \textsubscript{$\pm$ 0.199} & 8.258 \textsubscript{$\pm$ 0.150} & 0.014 \textsubscript{$\pm$ 0.000} & 0.124 \textsubscript{$\pm$ 0.001} \\

        $\sigma_\text{CFG}=5$ (Ours)  
        & 3.072 \textsubscript{$\pm$ 0.199} & 9.220 \textsubscript{$\pm$ 0.151} & 0.300 \textsubscript{$\pm$ 0.007} & 0.481 \textsubscript{$\pm$ 0.004} & 0.607 \textsubscript{$\pm$ 0.008} 
        & 4.272 \textsubscript{$\pm$ 0.058} & 3.238 \textsubscript{$\pm$ 0.430} & 8.226 \textsubscript{$\pm$ 0.079} & 0.015 \textsubscript{$\pm$ 0.001} & 0.125 \textsubscript{$\pm$ 0.001} \\
        
        $\sigma_\text{CFG}=6$ 
        & 3.588 \textsubscript{$\pm$ 0.148} & 9.197 \textsubscript{$\pm$ 0.071} & 0.304 \textsubscript{$\pm$ 0.001} & 0.486 \textsubscript{$\pm$ 0.002} & 0.610 \textsubscript{$\pm$ 0.010} 
        & 4.141 \textsubscript{$\pm$ 0.016} & 4.015 \textsubscript{$\pm$ 0.282} & 8.279 \textsubscript{$\pm$ 0.090} & 0.015 \textsubscript{$\pm$ 0.000} & 0.125 \textsubscript{$\pm$ 0.000} \\
        
        $\sigma_\text{CFG}=7$ 
        & 3.702 \textsubscript{$\pm$ 0.118} & 9.195 \textsubscript{$\pm$ 0.109} & 0.306 \textsubscript{$\pm$ 0.013} & 0.489 \textsubscript{$\pm$ 0.009} & 0.608 \textsubscript{$\pm$ 0.010} 
        & 4.094 \textsubscript{$\pm$ 0.037} & 4.365 \textsubscript{$\pm$ 0.369} & 8.323 \textsubscript{$\pm$ 0.138} & 0.016 \textsubscript{$\pm$ 0.000} & 0.127 \textsubscript{$\pm$ 0.001} \\
        \midrule
        
        $\text{Steps}=3$ 
        & 3.755 \textsubscript{$\pm$ 0.057} & 9.111 \textsubscript{$\pm$ 0.164} & 0.309 \textsubscript{$\pm$ 0.010} & 0.481 \textsubscript{$\pm$ 0.011} & 0.608 \textsubscript{$\pm$ 0.007} 
        & 4.078 \textsubscript{$\pm$ 0.017} & 4.786 \textsubscript{$\pm$ 0.193} & 8.542 \textsubscript{$\pm$ 0.144} & 0.017 \textsubscript{$\pm$ 0.000} & 0.384 \textsubscript{$\pm$ 0.001} \\

        $\text{Steps}=5$ (Ours)
        & 3.072 \textsubscript{$\pm$ 0.199} & 9.220 \textsubscript{$\pm$ 0.151} & 0.300 \textsubscript{$\pm$ 0.007} & 0.481 \textsubscript{$\pm$ 0.004} & 0.607 \textsubscript{$\pm$ 0.008} 
        & 4.272 \textsubscript{$\pm$ 0.058} & 3.238 \textsubscript{$\pm$ 0.430} & 8.226 \textsubscript{$\pm$ 0.079} & 0.015 \textsubscript{$\pm$ 0.001} & 0.125 \textsubscript{$\pm$ 0.001} \\
        
        $\text{Steps}=7$ 
        & 3.451 \textsubscript{$\pm$ 0.047} & 9.230 \textsubscript{$\pm$ 0.078} & 0.308 \textsubscript{$\pm$ 0.014} & 0.483 \textsubscript{$\pm$ 0.007} & 0.616 \textsubscript{$\pm$ 0.006} 
        & 3.855 \textsubscript{$\pm$ 0.012} & 4.913 \textsubscript{$\pm$ 0.212} & 8.427 \textsubscript{$\pm$ 0.114} & 0.017 \textsubscript{$\pm$ 0.001} & 0.127 \textsubscript{$\pm$ 0.002} \\
        
        $\text{Steps}=10$ 
        & 3.640 \textsubscript{$\pm$ 0.072} & 8.951 \textsubscript{$\pm$ 0.106} & 0.317 \textsubscript{$\pm$ 0.008} & 0.497 \textsubscript{$\pm$ 0.007} & 0.623 \textsubscript{$\pm$ 0.007} 
        & 3.858 \textsubscript{$\pm$ 0.022} & 5.132 \textsubscript{$\pm$ 0.309} & 8.350 \textsubscript{$\pm$ 0.117} & 0.021 \textsubscript{$\pm$ 0.000} & 0.133 \textsubscript{$\pm$ 0.001} \\
        \midrule
        
        $\text{Rollout}=0$ 
        & 2.310 \textsubscript{$\pm$ 0.112} & 8.540 \textsubscript{$\pm$ 0.044} & 0.208 \textsubscript{$\pm$ 0.003} & 0.349 \textsubscript{$\pm$ 0.005} & 0.474 \textsubscript{$\pm$ 0.015} 
        & 5.936 \textsubscript{$\pm$ 0.015} & 1.129 \textsubscript{$\pm$ 0.159} & 8.087 \textsubscript{$\pm$ 0.214} & 0.011 \textsubscript{$\pm$ 0.000} & 0.123 \textsubscript{$\pm$ 0.001} \\
        
        $\text{Rollout}=0.4$ 
        & 2.461 \textsubscript{$\pm$ 0.089} & 8.831 \textsubscript{$\pm$ 0.074} & 0.262 \textsubscript{$\pm$ 0.010} & 0.419 \textsubscript{$\pm$ 0.017} & 0.543 \textsubscript{$\pm$ 0.008} 
        & 4.960 \textsubscript{$\pm$ 0.040} & 2.525 \textsubscript{$\pm$ 0.202} & 8.541 \textsubscript{$\pm$ 0.102} & 0.015 \textsubscript{$\pm$ 0.000} & 0.125 \textsubscript{$\pm$ 0.000} \\
        
        $\text{Rollout}=0.6$ 
        & 3.323 \textsubscript{$\pm$ 0.016} & 9.198 \textsubscript{$\pm$ 0.075} & 0.305 \textsubscript{$\pm$ 0.017} & 0.479 \textsubscript{$\pm$ 0.013} & 0.610 \textsubscript{$\pm$ 0.006} 
        & 4.176 \textsubscript{$\pm$ 0.045} & 3.537 \textsubscript{$\pm$ 0.201} & 8.386 \textsubscript{$\pm$ 0.074} & 0.013 \textsubscript{$\pm$ 0.001} & 0.124 \textsubscript{$\pm$ 0.001} \\

        $\text{Rollout}=0.8$ (Ours)
        & 3.072 \textsubscript{$\pm$ 0.199} & 9.220 \textsubscript{$\pm$ 0.151} & 0.300 \textsubscript{$\pm$ 0.007} & 0.481 \textsubscript{$\pm$ 0.004} & 0.607 \textsubscript{$\pm$ 0.008} 
        & 4.272 \textsubscript{$\pm$ 0.058} & 3.238 \textsubscript{$\pm$ 0.430} & 8.226 \textsubscript{$\pm$ 0.079} & 0.015 \textsubscript{$\pm$ 0.001} & 0.125 \textsubscript{$\pm$ 0.001} \\
        
        $\text{Rollout}=1$ 
        & 4.008 \textsubscript{$\pm$ 0.026} & 9.185 \textsubscript{$\pm$ 0.137} & 0.315 \textsubscript{$\pm$ 0.005} & 0.500 \textsubscript{$\pm$ 0.008} & 0.624 \textsubscript{$\pm$ 0.008} 
        & 3.847 \textsubscript{$\pm$ 0.006} & 4.363 \textsubscript{$\pm$ 0.253} & 8.192 \textsubscript{$\pm$ 0.062} & 0.018 \textsubscript{$\pm$ 0.001} & 0.129 \textsubscript{$\pm$ 0.003} \\
        \bottomrule
    \end{tabular}}
    \label{tab:ablation_study}
\end{table*}

\subsection{Additional Ablation Studies}
\label{app:exp:ablation}
As shown in Table~\ref{tab:ablation_study}, we conduct additional ablation studies to analyze the impact of key design choices in the motion generation module, including history length $T_{\text{history}}$, future length $T_{\text{future}}$, number of primitives $N_{\text{prim}}$, hidden dimension and layer count of LDM, guidance scale $\sigma_\text{CFG}$, number of denoising steps, and maximum rollout probability. Through a trade-off analysis considering distribution matching, semantic alignment, and motion smoothness, we ultimately select a set of parameters that achieves balanced performance across all metrics.

\begin{figure*}[t]
  \centering
  \includegraphics[width=0.9 \linewidth]{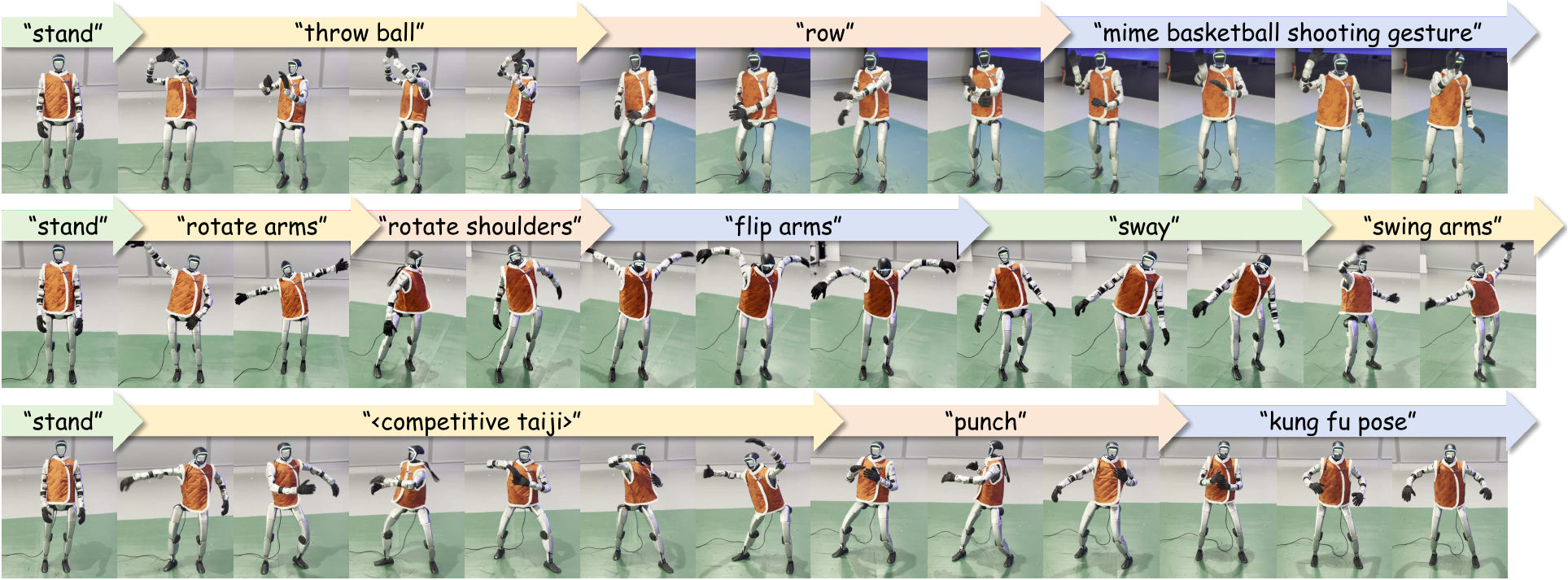}
  \caption{\textbf{Additional results of continuous diverse skill execution in the real robot.}}
  \label{fig:appendix_real_robot}
\end{figure*}
\subsection{Additional Real-World Results}
\label{app:exp:realworld}
Additional real-world robot results are shown in Fig.~\ref{fig:appendix_real_robot}.

\subsection{Computational Resources}
\label{app:exp:compute}

Model training is conducted on a server equipped with an NVIDIA A100-PCIE GPU with 40\,GB memory and an Intel Xeon Gold 6348 CPU operating at 2.60\,GHz. The training system runs Ubuntu~22.04.3~LTS.

For deployment, we use a separate workstation with an NVIDIA GeForce RTX~4090 GPU and a 13th Gen Intel Core i7-13700 CPU. The deployment environment runs Ubuntu~20.04.6~LTS.

\end{document}